\RequirePackage{amsthm}

 \documentclass[sn-nature]{sn-jnl}

\usepackage{graphicx}%
\usepackage{multirow}%
\usepackage{amsmath,amssymb,amsfonts}%
\usepackage{amsthm}%
\usepackage{mathrsfs}%
\usepackage[title]{appendix}%
\usepackage{xcolor, soul}%
\usepackage{textcomp}%
\usepackage{manyfoot}%
\usepackage{booktabs}%
\usepackage{algorithm}%
\usepackage{algorithmicx}%
\usepackage{algpseudocode}%
\usepackage{listings}%
\usepackage{lmodern} 
\usepackage{anyfontsize} 
\usepackage{adjustbox}

\usepackage{graphicx}
\usepackage{caption}
\usepackage{subcaption}
\usepackage{parskip}
\usepackage{comment}
\usepackage{array}
\usepackage{color,soul}
\usepackage{makecell}
\usepackage{float}

\newcolumntype{C}[1]{>{\centering\let\newline\\\arraybackslash\hspace{0pt}}m{#1}}

\newcommand{\ignore}[1]{}

\theoremstyle{thmstyleone}%
\theoremstyle{thmstyletwo}%

\theoremstyle{thmstylethree}%

\begin{document}

\title[Article Title]{Closed-Form Feedback-Free Learning with Forward Projection}


\author*[1]{\fnm{Robert} \sur{O'Shea}}\email{k1930297@kcl.ac.uk}

\author*[1]{\fnm{Bipin} \sur{Rajendran}}\email{bipin.rajendran@kcl.ac.uk}

\affil*[1]{\orgdiv{Centre for Intelligent Information Processing Systems, Department of Engineering}, \orgname{King's College London}, \orgaddress{\street{Strand}, \city{London}, \postcode{WC2R 2LS}, \state{London}, \country{UK}}}


\abstract{State-of-the-art backpropagation-free learning methods employ local error feedback to direct iterative optimisation via gradient descent. Here, we examine the more restrictive setting where retrograde communication from neuronal outputs is unavailable for pre-synaptic weight optimisation. We propose \textbf{Forward Projection} (FP), a randomised closed-form training method requiring only a single forward pass over the dataset without retrograde communication. FP generates target values for pre-activation membrane potentials through randomised nonlinear projections of pre-synaptic inputs and labels. Local loss functions are optimised using closed-form regression without feedback from downstream layers. A key advantage is interpretability: membrane potentials in FP-trained networks encode information interpretable layer-wise as label predictions. Across several biomedical datasets, FP achieves generalisation comparable to gradient descent-based local learning methods while requiring only a single forward propagation step, yielding significant training speedup. In few-shot learning tasks, FP produces more generalisable models than backpropagation-optimised alternatives, with local interpretation functions successfully identifying clinically salient diagnostic features. 
}


\maketitle

\section*{Introduction}\label{sec1}

A core task in neural network training is synaptic ``credit assignment'' for error in downstream layers \cite{Lee2014, Lillicrap2016Nov, Lillicrap2020Jun}. Although backpropagation has been established as the standard approach to this problem, its biological plausibility has been questioned \cite{Crick1989Jan, Lillicrap2020Jun, Millidge2021Jul}. Backpropagation requires bidirectional synaptic communication, which is incompatible with the unidirectional transmission of neural action potentials \cite{Lillicrap2020Jun}. Consequently, backpropagation would require either symmetric neural connectivity or a parallel retrograde network for error feedback to earlier layers \cite{Burbank2012Mar, Lillicrap2020Jun, Akrout2019Dec}. The backward pass traverses layers in reverse order of activation, leading to temporal discordance between forward and backward operations and necessitating storage of hidden activations. Furthermore, gradient descent requires hidden neural activations to be differentiable throughout. To approximate backpropagation in a more biologically plausible way, approaches such as Predictive Coding \cite{Friston2008, Millidge2021Jul} and Difference Target Propagation \cite{Lee2014} seek to align forward neural activity with a backward network that mirrors the forward architecture, effectively implementing an inverse model.  
Various strategies have been proposed to address the credit assignment problem with reduced retrograde communication requirements. Auxiliary loss functions computed on the activations of individual layers have been proposed to shorten the backward pass \cite{Bengio2007Jan,Kaiser2020May, Frenkel2021Feb, nokland19a}. A prominent example is the layer-wise greedy optimisation method that aims to minimise the costs of backpropagation by applying local supervision (LS) at each layer \cite{Bengio2007Jan, Wang2020Oct}. Extensions of this approach include deep continuous local learning \cite{Kaiser2020May}  and single layer updating \cite{Tang2022}. Hebbian learning rules using asymmetric feedback weights have been explored, allowing for independent updates for forward and feedback pathways \cite{Amit2019}.

Recent approaches have also explored using two forward passes to facilitate communication between upstream and downstream neurons \cite{Hinton2022Dec,dellaferrera2023,Srinivasan2023Feb, Pau2023}. The ``Forward-Forward'' (FF) learning algorithm \cite{Hinton2022Dec} is an approach in which data and label hypotheses are combined as inputs, with optimisation seeking to upregulate neural response to correctly labelled inputs and subdue responses to spuriously labelled inputs. However, inference under the original Forward-Forward algorithm requires a forward pass for each hypothesised label, presenting issues for tasks with large label spaces \cite{Pau2023}. The ``PEPITA" algorithm employed a preliminary forward pass to predict a label, which is used to generate spurious data-label instances for training \cite{dellaferrera2023}. 

The central issue with local learning methods in deep neural networks is that the optimal activity of hidden neurons is unknown during training, preventing direct observation of local error. Pre-defining activity targets for hidden neurons allows for heuristic local optimization. However, target definition strategies are a topic of ongoing research \cite{Song2024Feb}. Techniques such as local-representation alignment and target propagation introduce target values for hidden activations using limited retrograde communication from downstream neurons \cite{Lee2014, Ororbia2019Jul, Ororbia2023Feb, Frenkel2021Feb, Song2024Feb}.
Alternatively, target activities can be set as fixed random label projections computed during the forward pass, thereby permitting direct measurement of the error during the forward pass \cite{Frenkel2021Feb}. However, this approach may lead to highly correlated neuronal activity -- a problem known as informational collapse \cite{Wang2020Oct}. Although additive noise permits maximal decorrelation of target potentials \cite{Lee2014}, it is uninformative with respect to the label, potentially impeding model fitting. Approaches such as random neural network features \cite{Rahimi2007} forgo optimisation of hidden nodes entirely, instead projecting inputs to a random high-dimensional non-linear feature space. However, random feature layers require exponential scaling with respect to the input dimension to support downstream learning \cite{Yehudai2019}. 

The ``Predictive Coding'' paradigm provides a perspective on the collective functionality of intermediate neurons, proposing that local neural learning processes optimise the prediction of pre-synaptic neural inputs, minimising the ``surprise'' of out-of-distribution stimuli \cite{Millidge2021Jul}. Approaches such as the ``Difference Target Propagation'' \cite{Lee2014} reframe network layers as a series of autoencoders, where intermediate activations represent a series of encodings, transitioning from input information to label information. The issue of informational collapse in locally supervised SGD-trained models has been addressed by optimising the retention of information on the pre-synaptic activity \cite{Wang2020Oct}. ``Prospective configuration'' of target activities reformulates learning under the presumption that idealised adjustments to neuronal activity should be generated via energy minimisation before synaptic adjustment \cite{Song2024Feb}. Synaptic weights may be modified to realise the predetermined activities in response to the given input stimuli \cite{Song2024Feb}, thereby transitioning from input information to label information through model layers. ``Local neural synchronisation" generates target neural activity for hidden layers by projecting neuronal activity onto periodic basis vectors representative of class labels \cite{Apolinario2024May}. 

Going beyond these approaches, we propose the use of random nonlinear projections of both pre-synaptic inputs and target labels to generate local target activities in the forward pass. The objective of this approach is to develop a neural network training algorithm that requires no retrograde communication. Closed-form regression techniques are applicable in this setting, permitting single-step layer weight computation without error feedback. Thus, weight parameters are determined without backward communication from neuronal outputs or downstream layers. Further advantages of this method include the direct interpretability of hidden neurons with respect to local label predictions and  stability in the few-shot setting.

\section*{Results}

We consider a dataset $\mathcal{D}=\{(\mathbf{x}_i, \mathbf{y}_i)\}_{i=1}^{N}$ from the joint distribution $(X,Y)$. The task is to learn a feed-forward neural network function mapping $X\rightarrow Y$. The model has $L$ layers, with dimensions $m_0,\ldots, m_L$, and activations $\mathbf{a}_0,\ldots, \mathbf{a}_L$, where $\mathbf{a}_0=\mathbf{x}$ and $\mathbf{a}_L=\mathbf{y}$. Each layer is equipped with weights $\mathbf{W}_l$ to generate membrane potentials $\mathbf{z}_l=\mathbf{a}_{l-1}\mathbf{W}_l$, and activation function $f_l:\mathbb R\rightarrow \mathbb R$ to generate neuronal outputs $\mathbf{a}_l=f_l(\mathbf{z}_l)$. The model prediction is defined as $\hat{\mathbf{y}}=f_{L}(\mathbf{W}_L(\ldots f_1(\mathbf{W}_{1}\mathbf{x}))))$. 

\subsection*{Forward Projection (FP)}
\begin{figure}
\centering
\includegraphics[width=0.9\textwidth]{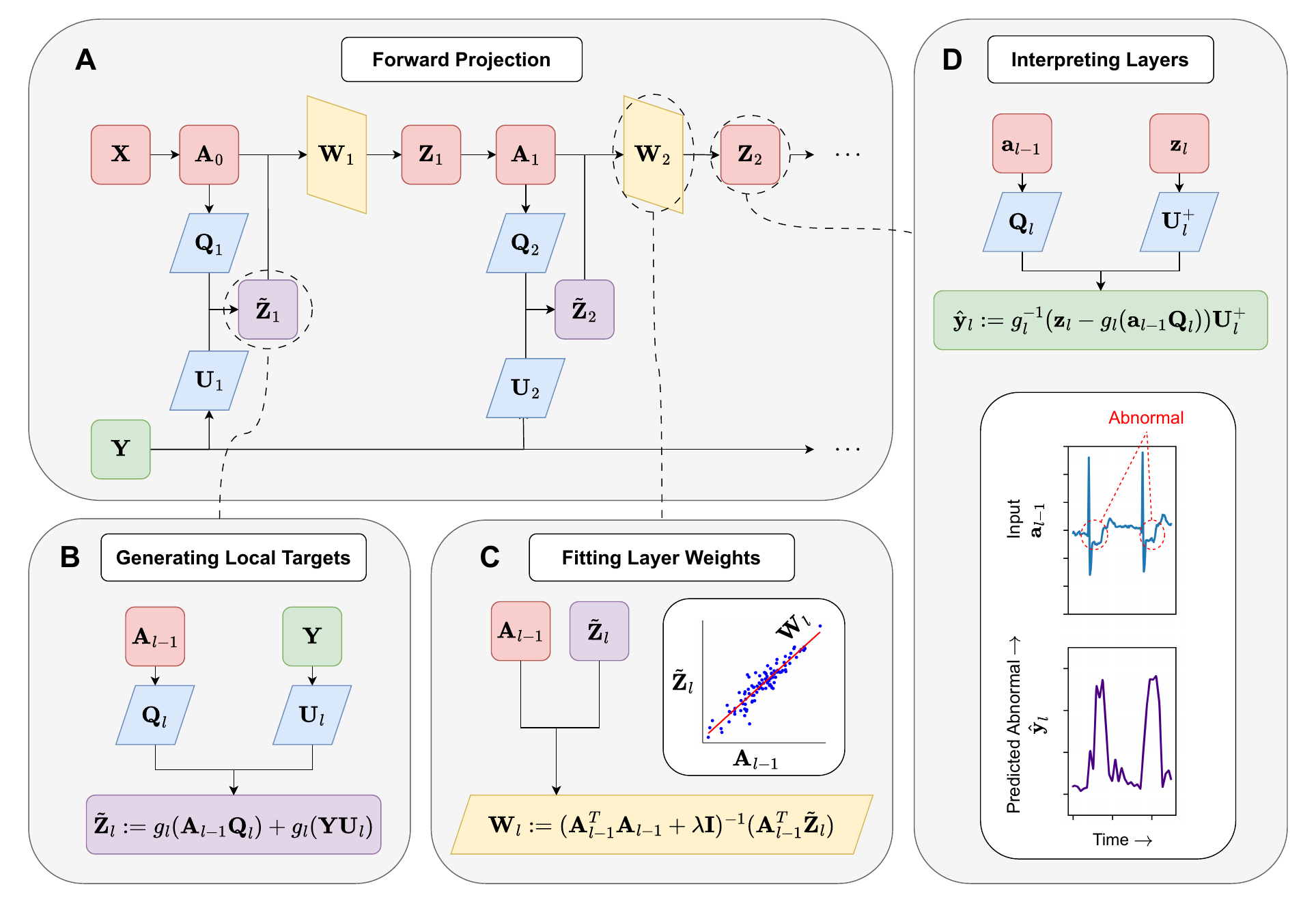}
\caption{\textcolor{black}{Graphical overview of forward projection.} A: Forward projection algorithm for fitting layer weights $\mathbf{W}_1, \ldots, \mathbf{W}_l$ to model labels $\mathbf{Y}$ from data $\mathbf{X}$. B: Procedure for generating the $l$-th layer target potentials $\tilde{\mathbf{Z}}_l$. Pre-synaptic inputs $\mathbf{A}_{l-1}$ and labels $\mathbf{Y}$ are projected with fixed matrices $\mathbf{Q}_l$ and $\mathbf{U}_l$, respectively, before applying non-linearity $g_l$. C: Optimising $\mathbf{W}_l$ to predict $\tilde{\mathbf{Z}}_l$ from $\mathbf{A}_{l-1}$ by ridge regression with penalty $\lambda$. D: Interpreting membrane potentials $\mathbf{z}_l$ as a local label prediction $\hat{\mathbf{y}}_l$ given pre-synaptic inputs $\mathbf{a}_{l-1}$ and projection matrices $\mathbf{Q}_l$ and $\mathbf{U}_l^+$, where $\mathbf{U}_l^+$ is the pseudo-inverse of $\mathbf{U}_l$.} \label{fig:fp_vis}
\end{figure}

To generate target membrane potentials for hidden neurons, we present the Forward Projection (FP) algorithm (Figure \ref{fig:fp_vis}-A). We propose to combine pre-synaptic inputs and labels using random non-linear projections to generate targets (Figure \ref{fig:fp_vis}-B). For each training sample, the target potential $\tilde{\mathbf{z}}_l\in \mathbb R^{1\times m_l} $ is generated from pre-synaptic inputs $\mathbf{a}_{l-1}\in \mathbb R^{1\times m_{l-1}}$ and labels $\mathbf{y}\in \mathbb R^{1\times m_L}$ using fixed random projection matrices $\mathbf{Q}_l\in \mathbb R^{m_{l-1}\times m_l }$ and $\mathbf{U}_l\in \mathbb R^{m_{L}\times m_l }$ such that \begin{equation}
 \tilde{\mathbf{z}}_l=g_l(\mathbf{a}_{l-1}\mathbf{Q}_l)+g_l(\mathbf{y}\mathbf{U}_l),\label{eq:fp target gen}
 \end{equation}
where $g_l: \mathbb R \rightarrow \mathbb R$ is an element-wise non-linear transformation. It is noted here that $g_l$ is not necessarily equal to $f_l$, the neural activation function. $\mathbf{Q}_l$ and $\mathbf{U}_l$ are fixed linear projections drawn from random Gaussian distributions, which are pre-defined before training. The target potential $\tilde{\mathbf{z}}_l$ allows a local auxiliary loss function to be defined with respect to the actual membrane potential, $\mathbf{z}_l$, realised during the forward pass. The randomised combination of pre-synaptic inputs with the target label is inspired by the high-dimensional computing paradigm \cite{Kanerva2009Jun}, where fixed random projections are employed to encode information from multiple vector inputs, approximately preserving relative distances according to the Johnson-Lindenstrauss lemma \cite{Kanerva2009Jun}. Accordingly, it is noted that $\tilde{\mathbf{z}}_l$ approximately encodes both $\mathbf{a}_{l-1}$ and $\mathbf{y}$ (see remark \ref{remark: interpretability} in SI). By generating target potentials in a forward manner, the mix of information encoded in each layer is expected to transition incrementally from predominantly input information in early layers to label information in later layers. We propose to define a local loss function $\mathcal{L}_l:=\lVert \mathbf{Z}_l - \tilde{\mathbf{Z}}_l\rVert$, which may be employed as an objective to optimise $\mathbf{W}_l$. Synaptic weights for each layer can be computed in a closed-form forward manner, using ridge regression (Figure \ref{fig:fp_vis}-C), such that 
\begin{equation}
\mathbf{W}_l:=(\mathbf{A}_{l-1}^\top\mathbf{A}_{l-1}+\lambda \mathbf{I})^{-1}(\mathbf{A}_{l-1}^\top\tilde{\mathbf{Z}}_l).\label{eq:fp fitting}  
\end{equation}
Here, $\mathbf{A}_{l-1}\in \mathbb R^{N\times m_{l-1}}$ and $\tilde{\mathbf{Z}}_l\in \mathbb R^{N\times m_l}$ are matrices of pre-synaptic activities and target potentials, respectively, collected over the $N$ samples in $\mathcal{D}$. $\lambda$ is a regularisation term and $\mathbf{I}$ is the identity matrix. Observe that  the $\mathbf{A}_{l-1}^\top\mathbf{A}_{l-1}$ and $\mathbf{A}_{l-1}^\top\tilde{\mathbf{Z}}_l$ terms in \eqref{eq:fp fitting} may be computed sequentially over instances $\{\mathbf{a}_{1, l-1}, \ldots,\mathbf{a}_{N, l-1}  \}\subset \mathbb R^{1\times m_{l-1}}$, and $\{\tilde{\mathbf{z}}_{1, l}, \ldots,\tilde{\mathbf{z}}_{N, l}  \} \subset \mathbb R^{1\times m_{l}}$ (see remark \ref{remark:seq fp comp} in SI). Therefore, memory requirements are independent of $N$, depending only on $m_{l-1}$ and $m_l$. \textcolor{black}{Note, weights are computed once only, after the single training epoch. }
The approach of encoding information on both inputs and labels in the hidden layers is inspired by the Predictive Coding and Target Propagation paradigms \cite{Millidge2021Jul, Lee2014, Bengio2007Jan}. However, unlike previous approaches, no backward communication is required from neuronal outputs to pre-synaptic inputs to achieve the FP fit. As a consequence of this fitting approach, neural membrane potentials in hidden layers of FP-trained models may be interpreted as label predictions (Figure \ref{fig:fp_vis}-C).

\subsection*{Forward Projection Performance}
Forward Projection was compared to other local learning methods, including random features (RF) \cite{Yehudai2019}, Local Supervision (LS) \cite{Bengio2007Jan}, and Forward-Forward (FF) \cite{Hinton2022Dec} in multiple tasks using equivalent model architectures (Table \ref{tbl:test perf main}). The Fashion MNIST (FMNIST) image classification task was modelled on a multi-layer perceptron (MLP). Forward Projection achieved higher test accuracy than other local learning methods in this task, approaching the performance of the backpropagation reference standard.
Two large-scale biomedical sequence modelling tasks were also evaluated with one-dimensional convolutional neural network (1D-CNN) architectures. The Promoters task required the identification of human non-TATA promoters, a class of gene promoter regions that increase transcription of DNA sequences \cite{Umarov2017Feb, Gresova2023Dec}. Forward projection yielded higher test performance than all other local learning methods in this task. 
The PTBXL-MI task \cite{Wagner2020May} required diagnosis of myocardial infarction (MI), a heart condition commonly known as ``heart attack", from 12-channel electrocardiogram (ECG) recordings. FP and LS performed comparably in this task. FP also outperformed all other methods in the optimisation of models with more complex neuronal activations such as modulo and polynomial activation functions (see remark \ref{challenging activations} in SI). Backpropagation performance benchmarks in PTBXL-MI and Promoters tasks were consistent with previous studies \cite{Anand2022May, Gresova2023Dec}. 
For the CIFAR10 classification task without data augmentation, two-dimensional convolutional neural networks (2D-CNNs) trained by FP underfitted, but still outperformed all other local learning methods, whereas standard backpropagation overfitted the model. FP was implemented in a transformer architecture to discriminate the first two classes of the CIFAR10 dataset (``CIFAR2"), significantly outperforming random features.

\subsubsection*{Alternative Feedback-free Approaches}
\begin{figure}
  \centering
  \includegraphics[width=0.9\linewidth]{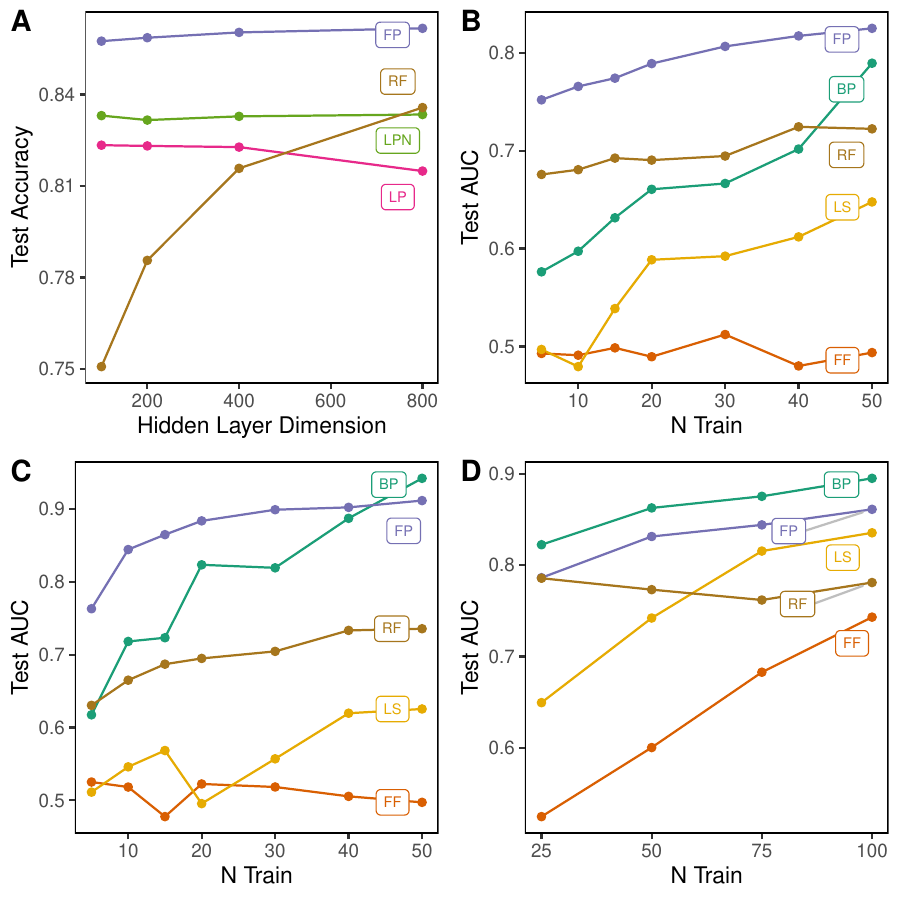}
   \caption{
\textcolor{black}{Performance of Forward Projection with backpropagation and local learning approaches.} A: Comparison of feedback-free fitting methods on FMNIST. MLP architectures had 1000 neurons in the first and second layers and 100, 200, 400 or 800 neurons in the final hidden layer. B-D: Test performance of few-shot trained 2D-CNN models.  Mean test AUC is reported over 50 few-shot training experiments. B: Chest X-ray (CXR) task. C: Optical Coherence Tomography (OCT) task. D: CIFAR2 task in which models were required to classify the first two classes (aeroplane and automobile). Models were fitted with $N\in\{5,10,15,20,30,40,50\}$ training samples from each class in CXR and OCT tasks and $N\in \{25,50,75,100\}$ samples per class for the CIFAR2 task. Predictive Coding and Difference Target Propagation are plotted separately in Supplementary Figure \ref{fig: fewshot extra methods}. BP: backpropagation; FF: Forward-Forward; FP: Forward Projection; LS: Local Supervision; RF: Random Features.}
  \label{fig: 2}
\end{figure}

To assess the value of the FP target generation function, we compared the performance of closed-form regression models fitted to targets generated by alternative functions, including simple label projection ($\tilde{\mathbf{Z}}_l:= \mathbf{yU}_l$) and label projection with additive noise $\left(\tilde{\mathbf{Z}}_l:= \mathbf{yU}_l + \mathbf{E}\right)$. To evaluate the capability of feedback-free training methods to handle information ``bottlenecks", they were applied to optimise MLP architectures for FMNIST classification. MLPs were generated with 1000 hidden neurons in the first and second layers ($m_1=m_2=1000$), and $m_3\in \{100, 200, 400, 800\}$ neurons in the final hidden layer. FP outperformed other feedback-free approaches (Figure \ref{fig: 2}-A). The performance of RF deteriorated in models with small penultimate layers, as relevant information was less likely to be represented by random projection \cite{Yehudai2019}. The performance of simple label projection deteriorated in models with large penultimate layers, a result that may be attributable to rank deficiency (see remark \ref{remark:lin dep} in SI). Noisy label projection maintained steady performance but at a lower level than FP. 

\subsection*{Few-Shot Learning}
``Few-shot" learning is a constrained learning scenario in which the number of data samples available for training is small. High-dimensional data, such as images, pose a challenge for few-shot learning, as many spurious features may exist. Thus, successful few-shot training methods must select generalisable features in the presence of these confounders. Few-shot learning was assessed in image classification tasks using a 2D-CNN architecture. The optical coherence tomography task (OCT) \cite{Kermany2018Feb} required that the model discriminate images of healthy retinas from those with choroid neovascularization, a pathology that affects the eye and manifests mainly as abnormal growth of blood vessels behind the retina. The paediatric chest X-ray (CXR) task \cite{Kermany2018Feb} required the model to discriminate between images of viral pneumonia, bacterial pneumonia, or healthy controls. The CIFAR2 task required the model to discriminate between the first two CIFAR10 classes (aeroplane and automobile). Few-shot training datasets were generated by subsampling, with $N\in \{5,10,15,20,30,40,50\}$ training examples from each class for OCT and CXR tasks, and  $N\in \{25,50,75,100\}$ training examples per class for the CIFAR2 task. Data augmentation was not employed. Model generalisability was assessed on all test data (CXR: $ N_{\text{test}}=431$; OCT: $N_{\text{test}}=327$; CIFAR2: $N_{\text{test}}=2000$).
Forward Projection-trained models demonstrated the greatest few-shot generalisability in  CXR (Figure \ref{fig: 2}-B) and OCT (Figure \ref{fig: 2}-C), outperforming all other methods, including backpropagation, in all the tasks with $N\le40$ training samples. In the CIFAR2 few-shot task, Forward Projection achieved the best performance of any local learning method in experiments with $N\in \{50, 75, 100\}$, being outperformed only by backpropagation. With as few as $N=10$ training samples, Forward Projection fitted discriminative models for the classification of OCT (test AUC: $84.5\pm 5.8$) and CXR (test AUC: $76.6 \pm 5.8$). CXR and OCT datasets highlighted two distinct vulnerabilities of backpropagation training in few-shot conditions. In the $N=5$ setting on OCT, backpropagation overfitted the training samples (Train AUC: $86.8 \pm 13.2$; Test AUC: $61.8 \pm 17.4$), as models integrated noise into decision functions. On the other hand, backpropagation failed to achieve adequate model fitting in the $N=10$ setting on CXR (Train AUC: $69.0 \pm 8.1$; Test AUC: $59.7 \pm 6.2$). Random features performed comparably to backpropagation in tasks with fewest training samples, but performance improved only slightly with larger sample sizes. Random Features models could not overfit within convolutional layers, as these contained no free parameters. However, Random Features had limited capacity to learn structural features in larger training samples. The improvement that Forward Projection provided over Random Features is therefore attributable to structural feature learning within hidden convolutional layers. Label Projection and Noisy Label Projection failed to converge in few-shot training (training AUC $\approx 0.50$). Neither Local Supervision nor Forward-Forward training matched the baseline generalisability of Random Features in OCT and CXR tasks - however Local Supervision achieved comparable test discrimination to Forward Projection in the CIFAR2 task when $N\ge75$. Predictive coding and Difference Target Propagation  yielded uninformative models in few-shot learning tasks (Figure \ref{fig: fewshot extra methods}). Few-shot performance is tabulated in Supplemental Table \ref{tab:fewshot_performance}.

\subsection*{Feature Interpretability} \label{section: interpretability}
\begin{figure}
  \centering
  \includegraphics[width=1\linewidth]{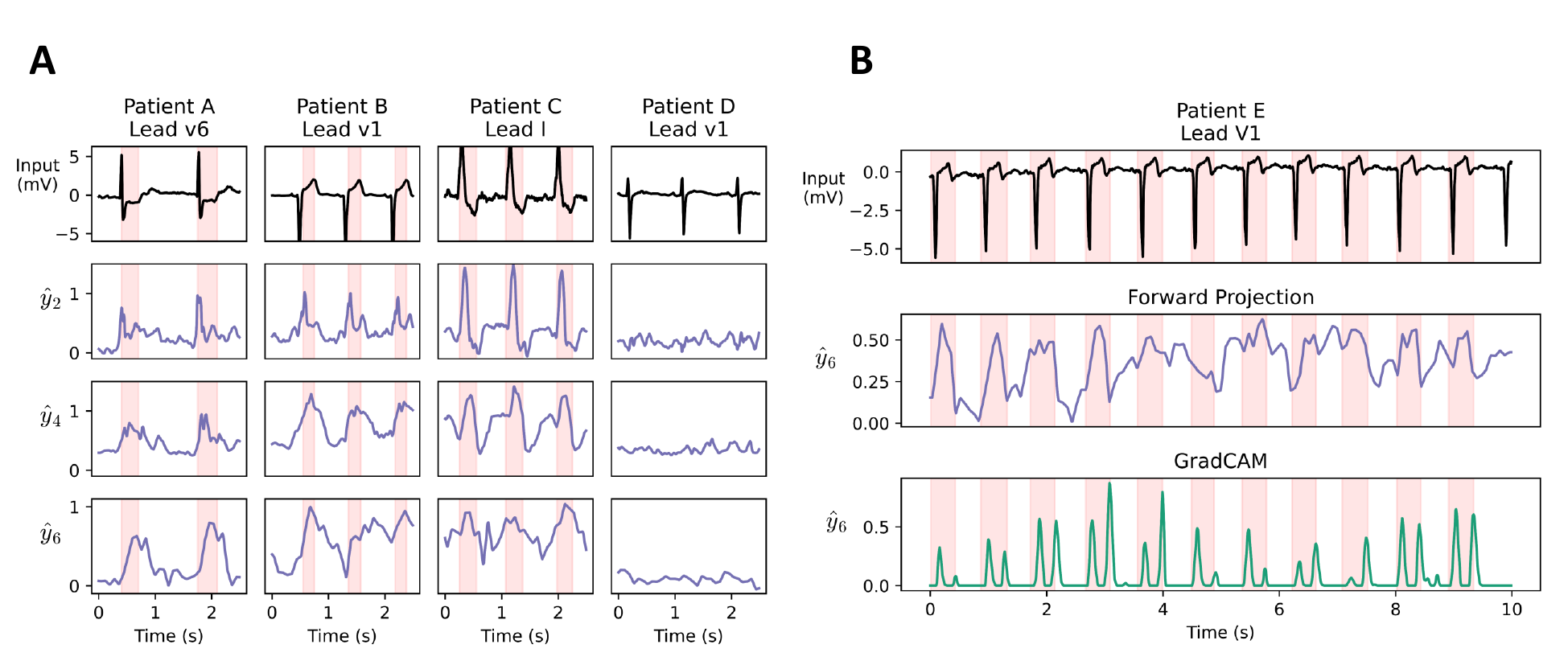}
  \caption{\textcolor{black}{Forward Projection layer interpretions for electrocardiogram analysis. }A: Visualisation of layer explanations over time in a 1D-convolutional neural network trained by Forward Projection to detect myocardial infarction (MI) in electrocardiograms (ECGs) from PTBXL data. Patients A, B, C, E (diagnosed MI) and patient D (no disease) were extracted from test data. Explanations were extracted from the second, fourth and sixth convolutional layers ($\hat{y}_2,\hat{y}_4,\hat{y}_6$) using equation \eqref{eq:layer explanation}. Explanations increase with MI features (highlighted in red), including ST-segment depression (Patient A), ST-segment elevation (Patient B) and QRS widening with T-wave inversion (Patient C). B: Comparison of Forward Projection with GradCAM.
  Top: ECG data from Patient E (diagnosed MI), showing ST-segment elevation. Middle: Sixth convolutional layer explanation ($\hat{y}_6$) from a model trained by Forward Projection. Below: GradCAM output for the sixth convolutional layer of a model trained by backpropagation.}
  \label{fig:Model Explainability conv1d}
\end{figure}

Explainability is a central issue with backpropagation-based learning, as relationships between hidden activations and model predictions may be non-monotonic, complicating the interpretation of hidden neural activities \cite{Saeed2023Mar}. An important advantage of Forward Projection is the interpretability of hidden neuron activity with respect to label predictions (see remark \ref{remark: interpretability} in SI). Assuming $\mathbf{z}_l\approx \tilde{\mathbf{z}}_l$, neural potentials may be interpreted as a local label prediction $\hat{\mathbf{y}}_l$ (Figure \ref{fig:fp_vis}-D), such that
\begin{equation}
  \hat{\mathbf{y}}_l:= g_l^{-1}(\tilde{\mathbf{z}}_l-g_l(\mathbf{a}_{l-1}\mathbf{Q}_l))\mathbf{U}_l^+.\label{eq:layer explanation}
\end{equation}
Here, $\mathbf{U}_l^+$ is the Moore-Penrose generalised inverse of the label projection matrix. Likewise, pre-synaptic inputs are encoded in neural pre-activation potentials, with an analogous reconstruction function (see remark \ref{remark: interpretability} in SI). It is noted that \eqref{eq:layer explanation} may be uninformative if $\frac{\lVert \mathbf{z}_l-\tilde{\mathbf{z}}_l \rVert}{\lVert \tilde{\mathbf{z}}_l \rVert}$ is large -- i.e., if $\mathbf{W}_l$ did not achieve a good fit. Measurement of local error in training data may provide insights into the reliability of \eqref{eq:layer explanation} during inference. Interpretation of pre-activation potentials $\mathbf{z}_l$ is simplified by selection of bijective functions for $g_l$ so that $g_l^{-1}$ exists everywhere. In practice, a surrogate approximation to the functional inverse was observed to suffice in our experiments; for example, $g_l^{-1}(\cdot)\approx\operatorname{tanh}(\cdot)$ was employed as a surrogate inverse for $g_l(\cdot)=\operatorname{sign}(\cdot)$. 
In our experiments, hidden neurons of models fitted with Forward Projection were interpretable as label predictions. In the FMNIST task, test accuracy of layer explanations was observed to improve between early layers and subsequent layers (Figure \ref{fig: mlp technical analysis}-A), demonstrating synergistic learning. In the PTBXL-MI task, applying the surrogate layer explanation function (described in equation \eqref{eq:layer explanation}) to the convolutional layers identified various clinically salient features for diagnosing myocardial infarction (MI). MI, a clinical condition characterised by damage to heart muscles due to poor blood flow, may manifest in ECG data with various electrophysical abnormalities. Consequently, the model must learn several distinct pathological waveform features, including elevation of the ``ST" segment, or inversion of the ``T"-wave. Figure \ref{fig:Model Explainability conv1d}-A shows the layer explanations as a function of time in four patients, three of whom were diagnosed with myocardial infarction. Patient A demonstrates ST-segment depression in lead II, which is temporally consistent with peaks in the model explanation functions at each layer. Likewise, the model explanation functions peak during ST-segment elevation in patient B and during QRS widening and T-wave inversion in Patient C. In contrast, the model explanation function is near zero in Patient D, who had normal ECG morphology. Forward Projection model explanations derived using \eqref{eq:layer explanation} from the sixth convolutional layer ($\hat{\mathbf{y}}_6$) were compared with GradCAM outputs of backpropagation-trained networks. The first 15 MI-positive test instances in the PTBXL dataset were annotated by a medical doctor to segment the diagnostically relevant subsequences. Layer explanations derived by Forward Projection achieved similar discrimination performance (AUC: $0.61\pm0.14$, AUPR: $0.52 \pm0.19$) to GradCAM (AUC: $0.59\pm0.12$, AUPR: $ 0.55\pm0.13$) (Figure \ref{fig:Model Explainability conv1d}-B).

\begin{figure}
  \centering
  \includegraphics[width=0.95\linewidth]{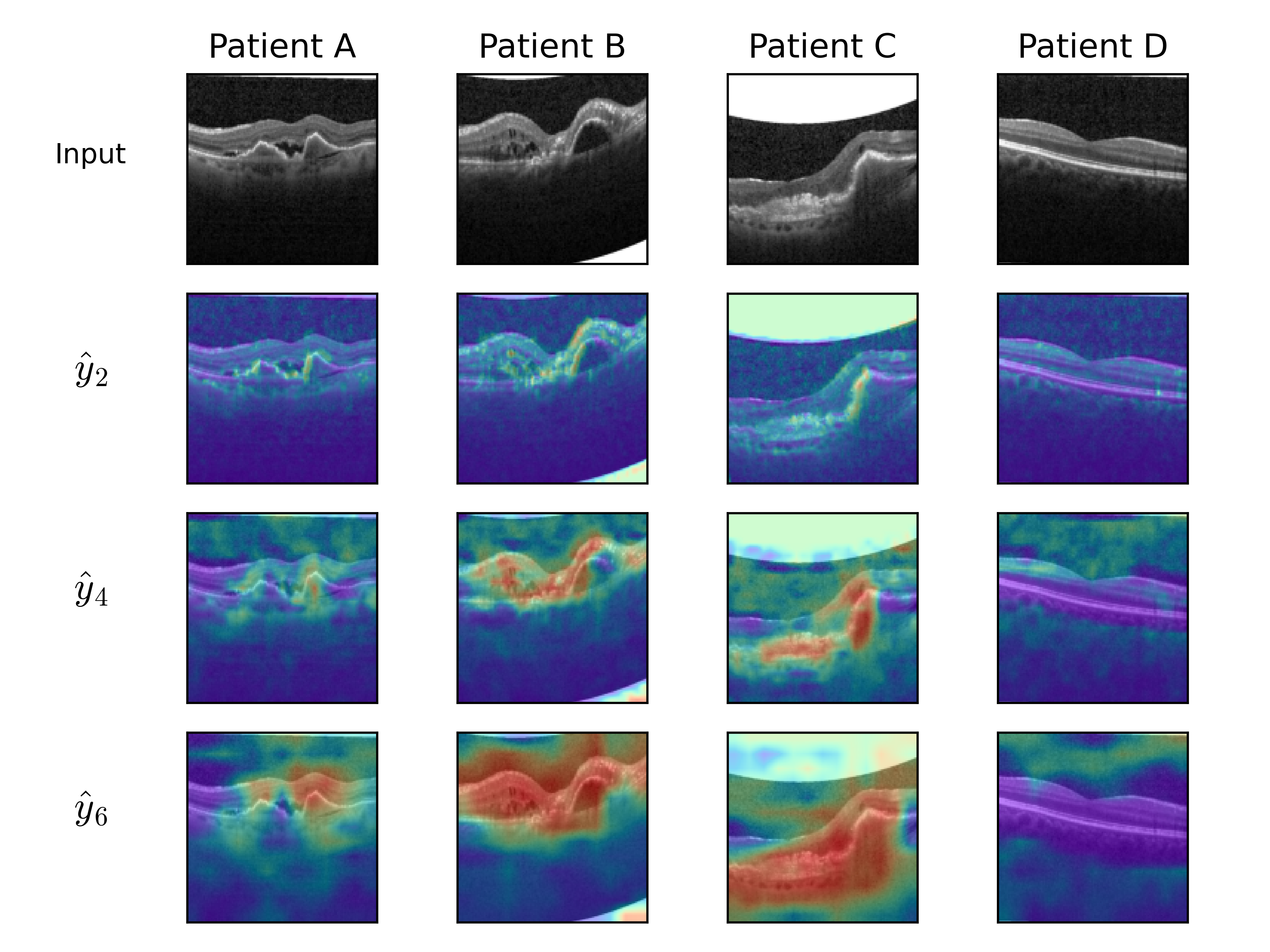}
  \caption{
Visualisation of layer explanations over space in 2D-CNNs trained by Forward Projection to detect choroid neovascularization (CNV) in the OCT task. Ensemble average of five models shown. Patients A-C (diagnosed CNV) and patient D (no disease) were extracted from test data. Explanations were extracted from the second, fourth and sixth convolutional layers ($\hat{y}_2,\hat{y}_4,\hat{y}_6$), using  \eqref{eq:layer explanation}. CNV heat-maps demonstrate high values (red) over CNV features, including retinal/subretinal fluid (Patients A-C) and hard exudates (Patient B), and fibrosis (Patient C), with low values (blue) over healthy retina (Patient D).}
  \label{fig:Model Explainability conv2d}
\end{figure}

Choroid neovascularization (CNV) is the growth of abnormal blood vessels behind the retina due to diseases such as age-related macular degeneration \cite{Wang2019Dec}. In OCT images, CNV may be represented by various image features, including hyper-reflective dots and detachment of the retinal pigment epithelium \cite{Metrangolo2021Jul}. In the OCT task, the model's layer explanation functions identified regions of interest related to CNV (Figure \ref{fig:Model Explainability conv2d}).
2D-CNN models trained with only 100 instances per class learned to localise fine-grained CNV features including retinal/subretinal fluid (Patients A-C), hard exudates (Patient B), and fibrosis (Patient C). 

\subsection*{Training Complexity }
We now analyse the complexity of training with the Forward Projection algorithm for a classification task by estimating the storage and computational requirements for a densely connected $m\times m$ hidden layer. We consider a dataset of $N$ training samples, with label dimension $m_L$. Note that FP model weights can be obtained in one pass over the $N$ training samples, while all other methods require each training sample to be fed to the network for several epochs (denoted $N_e$). The training procedure for each layer is presented in Supplementary Figure \ref{fig:local learning}.

The memory requirement for Forward Projection is $\mathcal{O}(m^2)$ for the layer weights, $\mathcal{O}(m^2)$ for the $\mathbf{Q}$ matrix, and $\mathcal{O}(mm_L)$ for the $\mathbf{U}$ matrix. As the $\mathbf{A}_{l-1}^\top\mathbf{A}_{l-1}$ and $\mathbf{A}_{l-1}^\top\tilde{\mathbf{Z}}_l$ terms in \eqref{eq:fp fitting} can be accumulated sequentially over data batches (See Remark \ref{remark:seq fp comp} in SI), two $m\times m$ matrices suffice for their storage, thereby avoiding storage of the $N\times m$ matrices $\mathbf{A}_{l-1}$ and $\tilde{\mathbf{Z}}_l$.
As with all methods, Forward Projection requires $\mathcal{O}(m^2)$ multiply-and-accumulate (MAC) operations to calculate the activations for the downstream layer in the forward pass. To generate target potentials during the forward pass, Forward Projection also requires $\mathcal{O}(m^2)$ additional MAC operations to project through $Q$ and $\mathcal{O}(mm_L)$ MAC operations to project through $\mathbf{U}$. 
Computation of the $\mathbf{A}_{l-1}^\top\mathbf{A}_{l-1}$ and $\mathbf{A}_{l-1}^\top\tilde{\mathbf{Z}}_l$ terms require $\mathcal O(Nm^2)$ MAC operations each. After all samples have been observed, computation of model weights using equation \eqref{eq:fp fitting} requires $\mathcal{O}(m^3)$ operations to invert the $\mathbf{A}_{l-1}^\top\mathbf{A}_{l-1}$ term and $\mathcal{O}(m^3)$ operations for matrix multiplication to complete the regression; however, this happens only once for each layer. Furthermore, Forward Projection requires no backward pass. 

For backpropagation, $\mathcal{O}(m^2)$ memory is required for storing layer weights, $\mathcal{O}(m^2)$ for accumulated gradients, and $\mathcal{O}(m)$ for activations. Here, the computation requirements scale as $\mathcal{O}(N_e m^2)$ MACs each for the forward pass, backward pass, and weight update calculations. Hence, in the typical setting where $m\gg m_L$, we note that Forward Projection and backpropagation have similar memory requirements, scaling as $\mathcal{O}(m^2)$. Notably, compute scales as $\mathcal{O}(m^3)$ for FP versus $\mathcal{O}(N_e m^2)$ for BP.
 For example, training a dense hidden layer with $1000$ inputs and $1000$ outputs on the FMNIST dataset ($N=60,000, m_L=10$) by Forward Projection requires a total of $1.2\times 10^{11}$ MAC operations for the forward pass and $1.2\times 10^{11}$ MAC operations for the weight update. Training the same layer via 100 epochs of backpropagation requires $6.0\times 10^{12}$ MAC operations for forward passes and $1.8\times 10^{13}$ operations for weight updates. Forward projection requires only a single training epoch, reducing overall computation time accordingly. Note that Local Supervision and Forward-Forward algorithms have similar computational and memory requirements as backpropagation, as illustrated in Table \ref{tab:complexity revised}. We also present the wall-clock time for full training of an MLP with $3\times1000$ hidden neurons on FMNIST, demonstrating a $66 \times$ speedup with Forward Projection in this example. Further timing results are provided in Table\ref{tab:timing performance} in SI. Computations were run using the Google Colab service with an NVIDIA T4 graphics processing unit. Training complexities of other methods are discussed in remark \ref{remark: train complexity other}. 
 
\section*{Discussion}  

We present the Forward Projection algorithm, which enables learning in a single pass over the dataset using random projections and closed-form optimisation. Compared to state-of-the-art local learning methods that require observing post-synaptic neuronal outputs to optimise synaptic weights via error-based gradient descent, Forward Projection operates under a stricter constraint, fitting weights using just the pre-synaptic neuronal activity and labels.

The target generation function proposed here for Forward Projection promotes input and label encoding in neural membrane potentials (see section \ref{remark: interpretability} in SI). Joint encoding of labels with pre-synaptic activity alleviates the degenerate neural activity which results from local modelling of simple label projections, whilst avoiding the introduction of uninformative noise (see section \ref{remark:lin dep} in SI). In this analysis, a simple non-linearity was employed for target generation. The utility of more complex nonlinearities for target generation is a subject for further research.

Explainability of neural network models is an important limitation in decision-critical fields such as biomedicine, where errors such as confounded decisions may lead to significant consequences \cite {Leming2022Jul}. FP-trained layers may be interpreted without downstream information, providing insight into model reasoning in hidden layers. FP interpretation yielded informative outputs in three model architectures, identifying clinically salient features in ECG sequences and OCT images. An important advantage of FP training is that saliency maps may be generated before downstream fitting, permitting on-the-fly inspection of intermediate layers for sufficiency or confounding. Thus, expert scrutiny of hidden layer performance may be conducted even before downstream architecture is finalised. 

Few-shot learning is ubiquitous in biological systems, which exhibit rapid neuronal adaptation to changing environments \cite{Benda2021Feb}. In our experiments, Forward Projection demonstrated clear performance advantages over backpropagation and local learning approaches in few-shot learning tasks, presenting a plausible method for learning new tasks rapidly. In this setting, convolutional features learned by Forward Projection yielded more generalisable models than backpropagation, which overfitted in some experiments and underfitted in others, even underperforming random features in some cases.  Forward Projection also maintained reasonable performance on activation functions which were untrainable by SGD-based methods (see section \ref{challenging activations} in SI). 

The closed-form Forward Projection fit is computable in a single pass over the data for each layer, presenting an opportunity to expedite training and reduce environmental footprint. The efficiency of FP training is attributable to a substantially different operational sequence from that employed in SGD-based approaches. Firstly, FP training collects a Gram matrix of pre-synaptic activity over a single epoch, which is subsequently inverted during a one-step weight matrix computation. Secondly, FP completes fitting for each layer before initialising successive layers. In contrast, SGD-based approaches fit all the layers in the network iteratively, using feedback from neuronal outputs and downstream layers. FP training requires no retrograde communication between neural output activations and pre-synaptic connections, enabling direct training on hardware with unidirectional synaptic and neuronal communication. This differs from current iterative local learning methods, which require backward communication from neuronal outputs to optimise pre-synaptic parameters (see remark  \ref{remark: local learning methods} in SI). Although FP addresses the feedback-free constraint inherent in biological learning systems, the biological plausibility of one-step learning via matrix inversion remains uncertain. Iterative fitting via gradient descent in sequential, batch-based learning offers a practical and scalable approach for aligning weights with locally generated targets, as defined in \eqref{eq:fp target gen} (see remark \ref{remark: iterative fp} in SI). Although closed-form optimisation remains a challenge for recurrent architectures, future research will explore integrating iterative training with random temporal convolutions to generalise the FP framework to dynamic models and expand its applicability to temporal learning tasks. This backpropagation-free strategy holds promise for biologically inspired computing systems, such as spiking neural networks, where non-differentiable activation functions preclude standard backpropagation \cite{Shen2024, Shen2025-rq, Shi2024}.  FP training may be applicable as a pre-training step to reduce the number of training epochs required for backpropagation.
In conclusion, FP is an efficient approach for neural network optimisation, employing techniques from randomised projective embedding and linear regression to fit weight matrices in one epoch using a single-step solution. Interpretability of hidden neurons in FP-trained models may be employed to improve the explainability of neural network predictions.

\section*{Methods}

Generalisability of machine learning methods to real-world datasets requires robustness to adverse modelling conditions such as class imbalance and noise, which often impede performance \cite{Shea2024Apr}. To assess the applicability and generalisability of Forward Projection and local learning methods in diverse real-world conditions, performance was evaluated in benchmark tasks from four biomedical domains described below.

\subsubsection*{PTBXL-MI}
The PTB-XL dataset contains 12-lead electrocardiography (ECG) recordings from 18,889 participants \cite{Wagner2020May, Goldberger2000Jun}. ECG recordings of ten-second duration and 100Hz sample rate were used in our experiments. The predictive task was to discriminate ECG recordings with normal waveform morphology ($N_{\text{train}}=6,451; N_{\text{test}}=721$) from those diagnosed as myocardial infarction ($N_{\text{train}}=2,707; N_{\text{test}}=268$) by a cardiologist. Data instances with uncertain diagnoses were excluded from this analysis. Following recommendations of the dataset authors who provided predefined participant-disjoint dataset splits, the tenth fold was held out for model testing \cite{Wagner2020May}. The first 15 MI-positive ECGs in the test set were annotated by a medical doctor to identify diagnostically relevant sections for quantitative explainability evaluation.

\subsubsection*{Promoters}
The Human Non-TATA Promoters dataset (``Promoters") was extracted from the GenomicBenchmarks repository \cite{Gresova2023Dec}. Data was originally published in \cite{Umarov2017Feb}. 36,131 nucleotide sequences of 251 bases each were analysed. Nucleotide sequences were converted to 4-channel one-hot vectors indicating adenine, cytosine, guanine and thymine. Indeterminate bases were represented with zero vectors. Models were required to classify the promoter functionality of the sequence as ``promoter" ($N_{\text{train}}=12,355; N_{\text{test}}=4,119$) or ``non-promoter" ($N_{\text{train}} =14,742;N_{\text{test}}=4,915$).

\subsubsection*{CXR}
The paediatric pneumonia chest X-ray dataset (``CXR") is a retrospective cohort of patients aged between one and five years, recorded in Guangzhou Women and Children’s Medical Center, Guangzhou, originally published in \cite{Kermany2018Feb}. Chest X-ray images were recorded as part of routine care during diagnostic workup for suspected lower respiratory tract infection. During data collection, clinicians screened the images for quality and excluded those with severe artefact or corruption. Images were annotated by two expert physicians. Local institutional review board approvals were obtained. Images were loaded in greyscale, rescaled to the $[0,1]$ intensity range and resized to $128\times 128$ pixels by bilinear interpolation. Models were required to classify images as ``normal" ($N_{\text{train}}=1,349; N_{\text{test}}=234$), ``viral pneumonia" ($N_{\text{train}}=2,538; N_{\text{test}}=242$), or ``bacterial pneumonia" ($N_{\text{train}}=1,345; N_{\text{test}}=148$).

\subsubsection*{OCT}
The optical coherence tomography dataset (``OCT") is a retrospective cohort of adult patients from five ophthalmology institutions in the USA and China recorded between 2013 and 2017 during routine care, originally published in \cite{Kermany2018Feb}. Images were initially annotated by local medical students, who had received OCT interpretation training. Subsequent annotation was performed by four ophthalmologists and two independent retinal specialists. Horizontal foveal cut images were available in portable network graphics image format. Images were loaded in greyscale, rescaled to the $[0,1]$ intensity range, and resized to $128\times 128$ pixels by bilinear interpolation. Models were required to classify images as either ``normal" ($N_{\text{train}}=2,926;N_{\text{test}}=149$) or ``choroid neovascularisation" ($N_{\text{train}}=791; N_{\text{test}}=178$).

\subsection*{Model Training}
The FMNIST dataset was modelled using an MLP with $3\times 1000$ hidden ReLU-activated neurons. Sequential datasets (PTBXL-MI and Promoters) were modelled by a 1D-CNN architecture of four convolutional blocks. Each convolutional block included two convolutional layers with kernel dimension $3$, and strides of 1 and 2 respectively. Convolutional layers in the $l$-th block had $32 \times 2^{l-1}$ filters. Convolutional implementation is detailed in Supplementary section \ref{remark: implementation fp conv}. Convolutional outputs were aggregated by global average pooling in the penultimate layer. For gradient-descent based learning algorithms, batch normalisation layers were included between convolutional blocks. CIFAR2 modelling in Table \ref{tbl:test perf main}  used a vision transformer architecture \cite{Dosovitskiy2021} operating on image patches of dimension $4\times4$, with a sequential stack of four multi-headed attention layers, each having 8 heads, embedding dimension 64, and MLP dimension 64. 2D-CNN architectures had kernel dimension $3\times 3$ and  the $l$-th convolutional block had  $16 \times 2^{l-1}$ filters (CXR and OCT),  $32 \times 2^{l-1}$ filters (CIFAR2), or  $64 \times 2^{l-1}$ filters (CIFAR10). The $\operatorname{sign}$ function was employed to generate target activations for Forward Projection models, such that
\begin{equation}
 \tilde{\mathbf{z}}_l=\operatorname{sign}(\mathbf{a}_{l-1}\mathbf{Q}_l)+\operatorname{sign}(\mathbf{yU}_l).\label{eq:fp_target_gen}
\end{equation}
Data augmentation was not performed in our experiments. Models were fitted using the PyTorch library. For SGD-based methods, early stopping was performed according to validation loss. Model weight initialisation was random, and optimisation was performed with the Adam optimiser with a learning rate of 0.001, training to minimise validation loss by early stopping with a patience of five epochs. In few-shot experiments, the learning rate was reduced to 0.0001 for SGD-based models and a patience of ten epochs was employed. It is acknowledged that direct comparison of gradient-descent based training with closed-form solutions is not strictly ``like-for-like". However, all efforts were made to ensure experimental conditions were otherwise equivalent. Models were fitted to minimise categorical or binary cross-entropy as appropriate. Implementations of Forward-Forward,  Local Supervision, Difference Target Propagation and Predictive Coding methods are detailed in supplementary section \ref{remark: implementation other methods}.

\subsection*{Explainability Analysis}
Model architectures for explainability analysis were equivalent to those used for the main experiments. Pre-activation potentials in hidden neurons were interpreted as local label predictions using \eqref{eq:layer explanation} with the following approximation
\begin{equation}
  \hat{\mathbf{y}}_l:=\tanh(\mathbf{z}_l-\operatorname{sign}(\mathbf{a}_{l-1}\mathbf{Q}_l))\mathbf{U}_l^+.\label{eq:fp_explainability1}
\end{equation}
Here, $\tanh(\cdot)$ is employed as a surrogate inverse for the $\operatorname{sign}$ function used to generate the target potentials.

\section*{Declarations}

 \subsection*{Ethics approval and consent to participate}
This study involves no new data collection or experiments on human or animal subjects. All clinical data used in this study was publicly available in previous publications \cite{Wagner2020May, Gresova2023Dec, Umarov2017Feb, Kermany2018Feb}. 
 
 \subsection*{Data Availability}
The PTBXL-MI dataset is available from PhysioNet \cite{Goldberger2000Jun}. The Promoters dataset is available from the GenomicBenchmarks library \cite{Gresova2023Dec} and the original publication \cite{Umarov2017Feb}. The CXR and OCT datasets are provided in \cite{Kermany2018Feb}. The pre-processed datasets and cross-validation folds used in this study are available in the Mendeley Database under accession code \href{https://data.mendeley.com/datasets/fb7xddyxs4/2}{fp\_datasets - Mendeley Data}  \cite{oshea25fp}.

\subsection*{Code Availability}
All code required to reproduce study findings is available from github.com/robertoshea/forward\_projection with digital object identifier (https://doi.org/10.5281/zenodo.17958941). 


\subsection*{Funding}
 This research was supported in part by the EPSRC Open Fellowship EP/X011356/1. 

 \subsection*{Acknowledgements}
\textcolor{black}{Study authors are grateful to Dr. Clément Ruah for valuable discussions}. 

\subsection*{Author Contributions}
\textbf{Robert O'Shea:} conceptualisation, methodology, formal analysis, software, investigation, visualisation, writing -- original draft preparation. \textbf{Bipin Rajendran:} conceptualisation, methodology, supervision, writing -- review \& editing, resources, project administration, funding acquisition.

 \subsection*{Competing Interests}
Robert O'Shea contributed to the initial draft of the manuscript whilst employed at King's College London. He commenced employment with Eli Lilly \& Company prior to the submission of the revised manuscript. His involvement in this project is entirely independent of his role at Eli Lilly \& Company and was not influenced by his employment there.

\clearpage
\section*{Tables}

\begin{table}[h!]
\tiny
\begin{tabular}{lllllllll}
\toprule
 &  & \multicolumn{7}{r}{} \\
 & Method & FP (ours) & RF & LS & FF & PC & DTP & BP (ref.)\\
Dataset & Metric &  &  &  &  &  &  &  \\
\midrule
\multirow[t]{2}{*}{FMNIST MLP}& AUC & 98.3 ± 0.0 & 98.0 ± 0.0 & 98.7 ± 0.1 & 98.5 ± 0.1 & 98.2 ± 0.1 & 97.8 ± 0.2 & 99.0 ± 0.0 \\
 & Acc & 86.3 ± 0.1 & 84.0 ± 0.2 & 83.1 ± 2.0 & 85.6 ± 0.8 & 83.0 ± 0.4 & 79.3 ± 2.0 & 88.4 ± 0.3 \\

\multirow[t]{2}{*}{Promoters 1D-CNN}& AUC & 88.7 ± 0.5 & 80.0 ± 1.1 & 86.6 ± 1.1 & 83.5 ± 0.7 & 79.7 ± 1.7 & 78.8 ± 0.8 & 94.1 ± 0.3 \\
 & Acc & 81.8 ± 0.5 & 72.2 ± 1.4 & 79.7 ± 1.2 & 75.3 ± 3.6 & 70.5 ± 3.6 & 69.3 ± 3.3 & 87.4 ± 0.7 \\

\multirow[t]{2}{*}{PTBXL-MI 1D-CNN}& AUC & 95.5 ± 0.5 & 94.7 ± 1.8 & 97.3 ± 0.3 & 89.3 ± 3.8 & 86.0 ± 0.9 & 86.4 ± 2.2 & 99.3 ± 0.0 \\
 & Acc & 86.5 ± 1.1 & 83.4 ± 2.1 & 86.1 ± 4.0 & 69.4 ± 8.2 & 64.2 ± 5.0 & 70.1 ± 5.8 & 94.9 ± 0.6 \\

\multirow[t]{2}{*}{CIFAR2 2d-ViT}& AUC & 91.5 ± 0.5 & 83.1 ± 0.5 &  &  &  &  & 96.1 ± 0.4 \\
 & Acc & 83.5 ± 0.5 & 75.9 ± 0.6 &  &  &  &  & 89.2 ± 0.7 \\

 \multirow[t]{2}{*}{CIFAR10 2D-CNN}& AUC 
& 85.1 ± 0.2& 81.4 ± 0.5& 87.7 ± 0.5& 57.9 ± 1.7& 74.5 ± 0.3& 64.5 ± 2.3&89.8 ± 0.7\\
 & Acc & 48.8 ± 0.3& 40.6 ± 1.0& 48.4 ± 1.1& 13.1 ± 1.9& 28.5 ± 0.9& 15.3 ± 3.0&54.8 ± 1.5\\
 
\bottomrule
\end{tabular}

  \caption{Test performance of various learning methods across different datasets. The FMNIST dataset was modelled with a multi-layer perceptron; Promoters and PTBXL-MI datasets were modelled using 1D convolutional neural networks (1D-CNNs); CIFAR2 (first two CIFAR10 classes only) was modelled using vision transformers and CIFAR10 was modelled with 2D-CNNs. BP: backpropagation; DTP: Difference Target Propagation; FF: Forward-Forward; FP: Forward Projection; LS: Local Supervision; PC: Predictive Coding; RF: Random Features.}
  \label{tbl:test perf main}
\end{table}

 \begin{table}[h!]
 \begin{tabular}{|p{1.3 cm}|p{1.15 cm}|l|l|p{1.7 cm}|p{1.1 cm}|p{1.42 cm}|p{1.4cm}|}\hline
  & \multicolumn{3}{|c|}{Compute}&\multicolumn{2}{|c|}{Memory} & \multicolumn{2}{|c|}{FMNIST Training}\\\hline 
\textbf{Method}& \textbf{Training epochs}&\textbf{Forward pass}& \textbf{Weight update} &\textbf{Model parameters} &\textbf{Weight update} &\textbf{Time (s)}
 &\textbf{Epochs} \\ \hline 
Backprop-agation& $N_e$
&$N_eNm^2$& $3N_eNm^2$&$m^2$&$m^2+2m$ &$22.43\pm4.8$& $11.6\pm2.7$\\ \hline 
Local Supervision& $N_e$
&$N_eN(m^2+mm_L)$& $2N_eN(m^2+mm_L)$&$m^2+mm_L$&$m^2+2m$ &$46.9 \pm 12.5$&$10.8\pm3.56$\\ \hline 
Forward-Forward& $N_e$
&$2N_eNm^2$& $4N_eNm^2$&$m^2$&$m^2+2m$&$172\pm 56$&$40.2\pm13.8$\\\hline
 Predictive Coding& $N_e$
& $N_eNm^2$& $3N_eNm^2$& $2m^2$& $2m^2+2m$& $150\pm61$&$25.4\pm10.2$\\\hline
 Difference Target Prop.& $N_e$
& $N_eNm^2$& $8N_eNm^2$& $2m^2$& $2m^2+4m$& $72.7\pm26.4$&$13.6\pm5.3$\\\hline
Forward Proj. (Ours)& $1$&$N(2m^2+mm_L)$& $2Nm^2+2m^3$&$2m^2+mm_L$&$2m^2$ &${0.34\pm 0.1}$ &$1$\\ \hline
\end{tabular}
\caption{Training complexity for a single hidden layer with $m$ inputs, $m$ outputs, given a label with dimension $m_L$. $N$: training sample size. $N_e$: epochs, all parameters except activations are per layer. For simplicity, we assume batch size $B=1$ and that predictive coding trains for $k=1$ iteration. Complexity is detailed in remark \ref{remark: train complexity other} in SI.}
\label{tab:complexity revised}
\end{table}

\section*{Figure Legends}

\textbf{Figure 1}. Graphical overview of forward projection. A: Forward projection algorithm for fitting layer weights $\mathbf{W}_1, \ldots, \mathbf{W}_l$ to model labels $\mathbf{Y}$ from data $\mathbf{X}$. B: Procedure for generating the $l$-th layer target potentials $\tilde{\mathbf{Z}}_l$. Pre-synaptic inputs $\mathbf{A}_{l-1}$ and labels $\mathbf{Y}$ are projected with fixed matrices $\mathbf{Q}_l$ and $\mathbf{U}_l$, respectively, before applying non-linearity $g_l$. C: Optimising $\mathbf{W}_l$ to predict $\tilde{\mathbf{Z}}_l$ from $\mathbf{A}_{l-1}$ by ridge regression with penalty $\lambda$. D: Interpreting membrane potentials $\mathbf{z}_l$ as a local label prediction $\hat{\mathbf{y}}_l$ given pre-synaptic inputs $\mathbf{a}_{l-1}$ and projection matrices $\mathbf{Q}_l$ and $\mathbf{U}_l^+$, where $\mathbf{U}_l^+$ is the pseudo-inverse of $\mathbf{U}_l$.

\textbf{Figure 2}. Performance of Forward Projection with backpropagation and local learning approaches. A: Comparison of feedback-free fitting methods on FMNIST. MLP architectures had 1000 neurons in the first and second layers and 100, 200, 400 or 800 neurons in the final hidden layer. B-D: Test performance of few-shot trained 2D-CNN models.  Mean test AUC is reported over 50 few-shot training experiments. B: Chest X-ray (CXR) task. C: Optical Coherence Tomography (OCT) task. D: CIFAR2 task in which models were required to classify the first two classes (aeroplane and automobile). Models were fitted with $N\in\{5,10,15,20,30,40,50\}$ training samples from each class in CXR and OCT tasks and $N\in \{25,50,75,100\}$ samples per class for the CIFAR2 task. Predictive Coding and Difference Target Propagation are plotted separately in Supplementary Figure \ref{fig: fewshot extra methods}. BP: backpropagation; FF: Forward-Forward; FP: Forward Projection; LS: Local Supervision; RF: Random Features.

\textbf{Figure 3}. Forward Projection layer interpretions for electrocardiogram analysis. A: Visualisation of layer explanations over time in a 1D-convolutional neural network trained by Forward Projection to detect myocardial infarction (MI) in electrocardiograms (ECGs) from PTBXL data. Patients A, B, C, E (diagnosed MI) and patient D (no disease) were extracted from test data. Explanations were extracted from the second, fourth and sixth convolutional layers ($\hat{y}_2,\hat{y}_4,\hat{y}_6$) using equation \eqref{eq:layer explanation}. Explanations increase with MI features (highlighted in red), including ST-segment depression (Patient A), ST-segment elevation (Patient B) and QRS widening with T-wave inversion (Patient C). B: Comparison of Forward Projection with GradCAM. Top: ECG data from Patient E (diagnosed MI), showing ST-segment elevation. Middle: Sixth convolutional layer explanation ($\hat{y}_6$) from a model trained by Forward Projection. Below: GradCAM output for the sixth convolutional layer of a model trained by backpropagation.

\textbf{Figure 4.} Visualisation of layer explanations over space in 2D-CNNs trained by Forward Projection to detect choroid neovascularization (CNV) in the OCT task. Ensemble average of five models shown. Patients A-C (diagnosed CNV) and patient D (no disease) were extracted from test data. Explanations were extracted from the second, fourth and sixth convolutional layers ($\hat{y}_2,\hat{y}_4,\hat{y}_6$), using  \eqref{eq:layer explanation}. CNV heat-maps demonstrate high values (red) over CNV features, including retinal/subretinal fluid (Patients A-C) and hard exudates (Patient B), and fibrosis (Patient C), with low values (blue) over healthy retina (Patient D).

\section*{Abbreviations}
1D-CNN: one-dimensional convolutional neural network\\
2D-CNN: two-dimensional convolutional neural network\\
BP: backpropagation\\
CNN: convolutional neural network\\
CNV: choroid neovascularisation\\
CXR: chest x-ray (dataset)\\
DTP: Difference Target Propagation\\
ECG: electrocardiography\\
FF: Forward-Forward algorithm\\
FMNIST: FashionMNIST dataset\\
FP: Forward Projection algorithm\\
LP: label projection\\
LPN: noisy label projection\\
LS: Local Supervision\\
MAC: multiply and accumulate\\
MI: myocardial infarction\\
MLP: multi-layer perceptron\\
OCT: optical coherence tomography (dataset)\\
PC: Predictive Coding\\
PTBXL: PTBXL dataset\\
RF: random features\\
SGD: stochastic gradient descent\\


\clearpage

\begin{appendices}

\section{Supplementary Information}\label{secA1}

\subsection{Local Learning Methods}\label{remark: local learning methods}
\begin{figure}[!htb]
  \centering
  \includegraphics[width=0.5\linewidth]{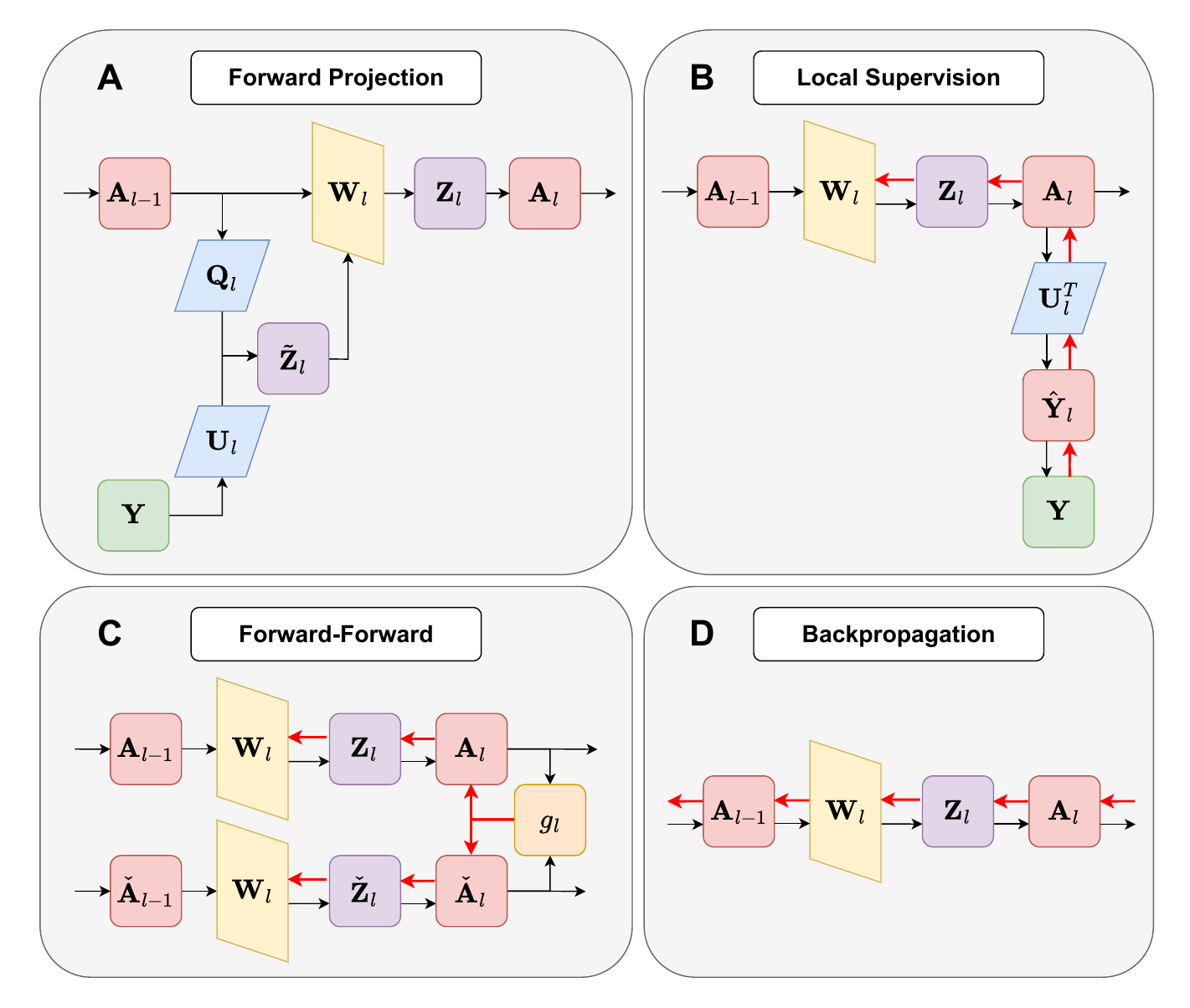}
  \caption{Training Procedures for Forward Projection, Local Supervision, Forward Forward and Backpropagation learning algorithms for the $l$-th hidden layer. A: Forward Projection generates target matrix $\hat{\mathbf{Z}}_l$ by projecting pre-synaptic inputs $\mathbf{A}_{l-1}$ by $\mathbf{Q}_l$ and labels $\mathbf{Y}$ by $\mathbf{U}_l$. Weights $\mathbf{W}_l$ are fitted by regression, generating membrane potential $\mathbf{Z}_l=\mathbf{A}_{l-1}\mathbf{W}_l$. B: In Local Supervision, an auxiliary prediction $\hat{\mathbf{Y}}_l$ is generated as a projection of the post-synaptic outputs $\mathbf{A}_l$, and $\mathbf{W}_l$ is updated by a short backward pass (red arrows). C: In Forward-Forward, ``positive" and ``negative" pre-synaptic activities, ${\mathbf{A}}_l$ and $\check{\mathbf{A}}_l$, are generated from true and spurious data-label pairs, respectively. $\mathbf{W}_l$ is updated to maximise positive activity whilst minimising negative activity. D: In backpropagation, $\mathbf{W}_l$ is updated along its gradient with respect to the backpropagated error.}
  \label{fig:local learning}
\end{figure}

Figure \ref{fig:local learning} illustrates training procedures and information flows in forward-projection, local-supervision, Forward-Forward and backpropagation algorithms.

\subsection{Convolutional Layer Implementation for FP}\label{remark: implementation fp conv}
We consider implementation of Forward Projection in a 2D convolutional layer over pre-synaptic input $\mathbf{a}_{l-1}\in \mathbb R ^{1\times m_{l-1}\times R\times C}$, having $m_{l-1}$ channels, and spatial dimensions $R\times C$ representing rows and columns. The $k_1\times k_2$ convolutional kernel positioned in row $r\in \{1,\ldots, R\}$ and column $c\in \{1,\ldots, C\}$ is represented in ``flattened" form as $\mathbf{a}_{l-1, r,c}\in \mathbb R ^{1\times (m_{l-1}k_1 k_2)}$. This pre-synaptic input is projected by the constant matrix $\mathbf{Q}_l\in \mathbb{R}^{(m_{l-1}k_1 k_2)\times m_l}$ to generate local target potentials $\tilde{\mathbf{z}}_{l,r,c}\in \mathbb R ^{1\times m_l}$, such that
\begin{equation}
 \tilde{\mathbf{z}}_{l,r,c}=g_l\left(\mathbf{a}_{l-1, r,c}\mathbf{Q}_l\right)+g_l\left(\mathbf{yU}_l\right).\label{eq:fp target gen conv2d}
 \end{equation}
 Weights were fitted such that
\begin{equation}
\mathbf{W}_l:=\left(\sum^R_{r=1}\sum^C_{c=1} \mathbf{A}_{l-1, r,c}^\top\mathbf{A}_{l-1, r,c}+\lambda \mathbf{I}\right)^{-1}\left(\sum^R_{r=1}\sum^C_{c=1} \mathbf{A}_{l-1, r,c}^\top\tilde{\mathbf{Z}}_{l,r,c}\right).\label{eq:fp fitting conv2d}  \end{equation}
Here, $\mathbf{A}_{l-1, r,c}\in \mathbb R^{N\times (m_{l-1}k_1k_2)}$ is a matrix representing the pre-synaptic input for the convolutional kernel in row $r$ and column $c$ of the $l$-th layer over all $N$ training instances. Likewise, $ \tilde{\mathbf{Z}}_{l,r,c}\in \mathbb R^{N\times m_l}$ contains target potentials for the kernel in row $r$ and column $c$ of the $l$-th layer over all $N$ training instances. In each task, the FP penalty parameter was fixed at $\lambda=10$ for hidden layers and $\lambda=1$ for output layers. To prevent numerical overflow due to large layers or training datasets, a scalar reduction factor $\tau\in (0,1]$ may be applied to downscale $\mathbf{A}^\top\mathbf{A}$, $\lambda$, and  $\mathbf{A}^\top\mathbf{Z}$ as necessary, such that:
\begin{equation}
\mathbf{W}_l:=\left(\sum^R_{r=1}\sum^C_{c=1} \tau\mathbf{A}_{l-1, r,c}^\top\mathbf{A}_{l-1, r,c}+ \tau\lambda \mathbf{I}\right)^{-1}\left(\sum^R_{r=1}\sum^C_{c=1}  \tau\mathbf{A}_{l-1, r,c}^\top\tilde{\mathbf{Z}}_{l,r,c}\right).\label{eq:fp fitting conv2d with tau}  \end{equation}

\subsection{Attention Layer Implementation for FP}\label{remark: implementation transformer}
We  consider the implementation of Forward Projection in a transformer layer \cite{Vaswani2017} over pre-synaptic input $\mathbf{a}_{l-1}\in \mathbb R ^{ R\times m_{l-1}}$, with  $R$ patches, each having $m_{l-1}$ channels. Each attention head of dimension $d\in \mathbb N$ contains three weight matrices for optimisation: $\mathbf{W}_l^{\operatorname{Query}}$,  $\mathbf{W}_l^{\operatorname{Key}}$ and $\mathbf{W}_l^{\operatorname{Value}}$, yielding intermediate activations $\mathbf{z}_l^{\operatorname{Query}}=\mathbf{a}_{l-1}\mathbf{W}_l^{\operatorname{Query}}$,   $\mathbf{z}_l^{\operatorname{Key}}=\mathbf{a}_{l-1}\mathbf{W}_l^{\operatorname{Key}}$, and $\mathbf{z}_l^{\operatorname{Value}}=\mathbf{a}_{l-1}\mathbf{W}_l^{\operatorname{Value}}$ respectively. The attention outputs $\mathbf{s}_l$ are then calculated as 
\begin{equation}
\mathbf{s}_l:={ \operatorname{Softmax}\left({\frac {\mathbf{z}_l^{\operatorname{Query}}\times \left(\mathbf{z}_l^{\operatorname{Key}}\right)^\top}{\sqrt {d}}}\right)\times \mathbf{z}_l^{\operatorname{Value}}}
\label{eq:transformer attention}
\end{equation}
Forward projection may be applied to generate target values for each of $\mathbf{z}_l^{\operatorname{Query}}, \mathbf{z}_l^{\operatorname{Key}}, \mathbf{z}_l^{\operatorname{Value}}$, using distinct data projection matrices  $\mathbf{Q}_l^{\operatorname{Query}}, \mathbf{Q}_l^{\operatorname{Key}}, \mathbf{Q}_l^{\operatorname{Value}}$, and a shared label projection matrix $\mathbf{U}_l$, such that for each patch, query matrix target potentials are given by
\begin{equation}
\tilde{\mathbf{z}}_{l,r}^{\operatorname{Query}}=g_l\left(\mathbf{a}_{l-1, r}\mathbf{Q}_l^{\operatorname{Query}}\right)+g_l\left(\mathbf{yU}_l\right).
\label{eq:fp target gen transformer query}
 \end{equation}
Thus, the query weight matrix is given by the penalised regression solution \cite{Golub2013-lw}. 
 Weights were fitted such that
\begin{equation}
\mathbf{W}_l^{\operatorname{Query}}:=\left(\sum^R_{r=1}\mathbf{A}_{l-1, r}^\top\mathbf{A}_{l-1, r}+\lambda \mathbf {I}\right)^{-1}\left(\sum^R_{r=1} \mathbf{A}_{l-1, r}^\top\tilde{\mathbf{Z}}_{l,r}^{\operatorname{Query}}\right).
\label{eq:fp fitting ViT} 
\end{equation}
$\mathbf{W}_l^{\operatorname{Key}}$ and $\mathbf{W}_l^{\operatorname{Value}}$, are fitted in the same manner - using $\mathbf{Q}_l^{\operatorname{Key}}$ and $\mathbf{Q}_l^{\operatorname{Value}}$, respectively, in place of $\mathbf{Q}_l^{\operatorname{Query}}$ in \eqref{eq:fp target gen transformer query}.

\subsection{Implementation of  local learning methods}\label{remark: implementation other methods}
Local Supervision was implemented by generating auxiliary predictions $\hat{\mathbf{y}}_1,\ldots, \hat{\mathbf{y}}_{L-1}$ from a set of fixed linear operators $\mathbf{U}_1, \ldots \mathbf{U}_{L-1}$ such that $\hat{\mathbf{y}}_l=f_L(\mathbf{a}_l\mathbf{U}_l^+)$. In each layer, $\mathbf{W}_l$ was optimised to minimise the auxiliary loss $\mathcal{L_l}(\hat{\mathbf{y}}_l, \mathbf{y})$ via gradient descent with two layers of backpropagation, such that
\begin{equation}  
\nabla_{\mathbf{W}_l}\mathcal{L}_l=\mathbf{a}^\top_{l-1}\left( 
f_l'(\mathbf{z}_l)\odot\left(f_L'(\mathbf{a}_l\mathbf{U}_l^+) \odot\nabla_{\hat{\mathbf{y}}_l}\mathcal{L}_l \right) (\mathbf{U}_l^+)^\top
\right).
\end{equation}
Here $ f_l'$ denotes the derivative of $f_l$. Local Supervision was implemented on convolutional layers by applying global-average pooling to generate a single vector representing mean neuronal activity over all convolutional windows, such that
\begin{equation}
\hat{\mathbf{y}}_l =f_L\left( \frac{1}{RC}\sum^R_{r=1}\sum^C_{c=1} \mathbf{a}_{l, r,c} \mathbf{U}_l^+\right). 
\end{equation}
The Forward-Forward algorithm \cite{Hinton2022Dec} was implemented by generating both ``positive" data-label pairs $[\mathbf{x}, \mathbf{y}]$ and ``negative" data pairs $[\mathbf{x}, \check{\mathbf{y}}]$ by concatenation. Positive activations were generated such that
\begin{equation}
\mathbf{a}^{\text{pos}}_l =f_l(\mathbf{W}_l\ldots f_1(\mathbf{W}_1 [\mathbf{x}, \mathbf{y}])\ldots ) . 
\end{equation}
Negative activations were generated with spurious labels $\check{y}\ne y$ such that
\begin{equation}
\mathbf{a}^{\text{neg}}_l =f_l(\mathbf{W}_l\ldots f_1(\mathbf{W}_1 [\mathbf{x}, \check{\mathbf{y}}])\ldots ).
\end{equation}
Thus, local auxiliary loss functions were computed using threshold hyperparameter $\theta=2$ and logistic sigmoid function $\sigma$, computing auxiliary loss $\mathcal{L}_l$ such that
\begin{equation}
\mathcal{L}_l:= \sigma (\lVert \mathbf{a}^{\text{neg}}_l\rVert^2_2-\theta )+\sigma (\theta- \lVert \mathbf{a}^{\text{pos}}_l\rVert^2_2).
\end{equation}
In MLP implementations, concatenation was implemented with supplementary input neurons to hold the label information as a one-hot vector. In convolutional implementations, the label was concatenated in the channel dimension, such that each channel indicated a single class, with a constant value over all kernel positions. 
Predictive Coding \cite{Friston2008, Millidge2021Jul} was implemented using inverse layers $\mathbf{W}_l^{\text{back}}\in \mathbb{R}^{m_l\times m_{l-1}}$ for each weight matrix $\mathbf{W}_l$, to estimate the pre-synaptic inputs from the post-synaptic outputs, such that
\begin{equation}
    \hat{\mathbf{a}}_{l-1} := f_l^{\text{back}}\left( \mathbf{a}_{l}\mathbf{W}_l^{\text{back}}\right) \label{eq:pc retro pred}
\end{equation}
The local loss is given by the reconstruction error
\begin{equation}
\mathcal{L}_l:= \lVert\mathbf{a}_{l-1} -  \hat{\mathbf{a}}_{l-1}   \rVert_2^2\label{eq:pred coding loss}
\end{equation}
Similarly, Difference Target Propagation \cite{Lee2014} aims to reconstruct presynaptic inputs, such that
\begin{equation}
    \hat{\mathbf{a}}_{l-1} := \mathbf{a}_{l-1}-f_l^{\text{back}}\left( \mathbf{a}_{l}\mathbf{W}_l^{\text{back}}\right)+f_l^{\text{back}}\left( \hat{\mathbf{a}}_{l}\mathbf{W}_l^{\text{back}}\right)
\end{equation}
In our implementations, we set $f_l^{\text{back}}=\operatorname{ReLU}$ in both Predictive Coding and Difference Target Propagation models.

\subsection{Analysis of label projection approaches}\label{remark:lin dep}
We now consider the necessity of pre-synaptic input projections, which distinguish Forward Projection from alternative feedback-free approaches such as simple label projection and noisy label projection. The objective is to generate a weight matrix $\mathbf{W}_l\in \mathbb R^{m_{l-1}\times m_l}$, given pre-synaptic inputs, $\mathbf{A}_{l-1}\in \mathbb R ^{N\times m_{l-1}}$, and label matrix $\mathbf{Y}\in \mathbb R^{N\times m_L}$.  As described previously, given some target potentials $\tilde{\mathbf{Z}}$, weights will be fitted by equation \eqref{eq:fp fitting}, such that
\begin{equation}
\mathbf{W}_l = (\mathbf{A}_{l-1}^\top\mathbf{A}_{l-1}+\lambda \textbf {I})^{-1}\mathbf{A}_{l-1}^\top\tilde{\mathbf{Z}}.
\end{equation}
We let $\operatorname{rank}(\tilde{\mathbf{Z}})$ denote the column rank of $\tilde{\mathbf{Z}}$ -- the maximum number of linearly independent column vectors in its column space. Since $\mathbf{W}_l$ is a linear function of $\tilde{\mathbf{Z}}$, we have $\operatorname{rank}(\mathbf{W}_l)\le \operatorname{rank}(\tilde{\mathbf{Z}})$.

We first consider target generation by simple label projection, such that $\tilde{\mathbf{Z}}_l:=\mathbf{YU}_l$. Here we have $\operatorname{rank}(\tilde{\mathbf{Z}})\le \operatorname{rank}(\mathbf{Y})\le m_L$. Where $m_l\gg m_L$, this implies severe degeneracy of $\mathbf{W}_l$, predisposing to correlated or redundant neuronal activities in $\mathbf{A}_l$. We next consider target generation by noisy label projection, using random Gaussian noise matrix $ \mathbf{E} \in \mathbb R^{N\times m_l}$ to perturb targets such that $\tilde{\mathbf{Z}}_l:=\mathbf{Y}\mathbf{U}_l + \mathbf{E}$. 
In this case, we have
\begin{equation}
  \operatorname{rank}(\tilde{\mathbf{Z}}_l)\le \operatorname{min}\left(m_l,  \left(\operatorname{rank}(\mathbf{Y})+ \operatorname{rank}(\mathbf{E})\right)\right) =m_l,
\end{equation}
However, as $\mathbf{E}$ is random and independent of $\mathbf{Y}$, this perturbation strategy may adversely affect label modelling of $\mathbf{Y}$. By adding projections of label and pre-synaptic inputs, such that $\tilde{\mathbf{Z}}_l:= \mathbf{A}_{l-1}\mathbf{Q}_l+ \mathbf{Y}\mathbf{U}_l$, we improve the upper bound to\begin{equation}
  \operatorname{rank}(\tilde{\mathbf{Z}}_l)\le \operatorname{min}\left(m_l, \left(\operatorname{rank}(\mathbf{A}_{l-1}) + \operatorname{rank}(\mathbf{Y})\right)\right) \le \operatorname{min}\left(m_l, (m_{l-1}+m_L)\right),
\end{equation}
without fitting $\mathbf{W}_l$ to noise. Accordingly, the targets generated by our proposed method $\tilde{\mathbf{Z}}_l:= g_l(\mathbf{a}_{l-1}\mathbf{Q}_l)+ g_l(\mathbf{Y}\mathbf{U}_l)$ may also have full rank.

\subsection{Computing the Forward Projection estimator}\label{remark:seq fp comp} 
Neural network models often manage storage demands by utilizing data mini-batches, allowing them to train on arbitrarily large datasets without a corresponding increase in memory requirements. Although the Forward Projection estimator described in equation \eqref{eq:fp fitting} includes large matrix terms $\mathbf{A}_{l-1}\in \mathbb R^{N\times m_{l-1}}$ and $\tilde{\mathbf{Z}}_l\in \mathbb R^{N\times m_{l}}$, these do not need to be loaded into memory. Assume $\mathbf{A}_{l-1}\in \mathbb R^{N\times m_{l-1}}$ and $\mathbf{Y}\in \mathbb R^{N\times m_{L}}$ are available as a set of row-vectors $\{\mathbf{a}_{i,l-1} \}^N_{i=1}\subset \mathbb R^{1\times m_{l-1}}$ and $\{\mathbf{y}_{i} \}^N_{i=1}\subset \mathbb R^{1\times m_{L}}$. 
The Gramian matrix is the sum of the row outer products, such that \begin{equation}
\mathbf{A}_{l-1}^\top\mathbf{A}_{l-1}=\sum^N_{i=1}\mathbf{a}_{i,l-1}^\top \mathbf{a}_{i,l-1}. \label{eq: aa sum}
\end{equation}
Therefore, the $\mathbf{A}_{l-1}^\top\mathbf{A}_{l-1}\in \mathbb R^{m_{l-1}\times m_{l-1}}$ term may be accumulated sequentially over instances $\{\mathbf{a}_{1, l-1}, \ldots,\mathbf{a}_{N, l-1}  \}$, with $\mathcal O(m_{l-1}^2)$ memory and $\mathcal O(Nm_{l-1}^2)$ compute, such that
\begin{equation}
(\mathbf{A}_{l-1}^\top\mathbf{A}_{l-1})^{(i)}=(\mathbf{A}_{l-1}^\top\mathbf{A}_{l-1})^{(i-1)} +\mathbf{a}_{i,l-1}^\top \mathbf{a}_{i,l-1}, \label{eq: aa}
\end{equation}
where $(\mathbf{A}_{l-1}^\top\mathbf{A}_{l-1})^{(0)}$ is the zero matrix of dimension $m_{l-1}\times m_{l-1}$. We may generate the corresponding target potentials $\{\tilde{\mathbf{z}}_{i,l} \}^N_{i=1}\subset \mathbb R^{1\times m_l}$, using predetermined matrices $\mathbf{Q}_l \in\mathbb R^{m_{l-1}\times m_l}$ and $\mathbf{U}_l \in\mathbb R^{m_{L}\times m_l}$according to \eqref{eq:fp target gen}. Thus, $\mathbf{A}_{l-1}^\top\tilde{\mathbf{Z}}_l\in \mathbb R^{m_{l-1}\times m_l}$ may also be accumulated sequentially such that
\begin{equation}
\mathbf{A}_{l-1}^\top\tilde{\mathbf{Z}}_l=\sum^N_{i=1}\mathbf{a}_{i,l-1}^\top \tilde{\mathbf{z}}_{i,l}, \label{eq: az}
\end{equation}
requiring $\mathcal O(m_{l-1}m_l)$ memory and $\mathcal O(Nm_{l-1}m_l)$ compute. Addition of the $\lambda\mathbf{I}$ regularisation term has negligible $\mathcal{O}(m_{l-1})$ overhead. Following sequential computation of equations \eqref{eq: aa} and \eqref{eq: az}, equation \eqref{eq:fp fitting} may be computed with one order $\mathcal{O}(m_{l-1}^3)$ matrix inversion and one order $\mathcal{O}(m_{l-1}^2m_l)$ matrix multiplication, yielding:
\begin{equation}
\mathbf{W}_l=(\mathbf{A}_{l-1}^\top\mathbf{A}_{l-1}+\lambda \mathbf{I})^{-1}(\mathbf{A}_{l-1}^\top\tilde{\mathbf{Z}}_l).\label{eq: fitting 3}
\end{equation}

\subsection{Information Encoding and Interpretability}\label{remark: interpretability}
We now consider the information encoding properties of Forward Projection training. Assuming $g_l^{-1}$ exists everywhere, equation \eqref{eq:fp target gen} relates labels $\mathbf{y}\in \mathbb R^{1\times m_L}$ to target potentials $\tilde{\mathbf{z}}_l\in\mathbb R^{1\times m_l}$, given presynaptic inputs $\mathbf{a}_{l-1}\in \mathbb R^{1\times m_{l-1}}$ via projection matrices $\mathbf{Q}_l\in \mathbb R^{m_{l-1}\times m_l}$ and $\mathbf{U}_l\in \mathbb R^{m_{L}\times m_l}$, such that
\begin{equation}
  \mathbf{y}\mathbf{U}_l=g_l^{-1}\left(\tilde{\mathbf{z}}_l-g_l(\mathbf{a}_{l-1}\mathbf{Q}_l)\right).
\end{equation}
Letting $\mathbf{U}_l^+$ denote the Moore-Penrose inverse, and assuming that $\mathbf{U}_l\mathbf{U}_l^+=\mathbf{I}$, we may reconstruct $\mathbf{y}$, such that
\begin{equation}
  \mathbf{y}= g_l^{-1}\left(\tilde{\mathbf{z}}_l-g_l(\mathbf{a}_{l-1}\mathbf{Q}_l)\right)\mathbf{U}_l^+.
\end{equation}
Assuming realised neuron pre-activation potentials $ \mathbf{z}_l$ are a good approximation of the target potentials, i.e., $\mathbf{z}_l\approx \tilde{\mathbf{z}}_l$, we may estimate the label from the hidden layer, such that
\begin{equation}
  \hat{\mathbf{y}}_l := g_l^{-1}\left(\mathbf{z}_l-g_l(\mathbf{a}_{l-1}\mathbf{Q}_l)\right)\mathbf{U}_l^+.\label{eq: fp label estimation}
\end{equation}
Thus, neural potentials are interpretable layer-wise as label predictions. Assuming  $\mathbf{Q}_l\mathbf{Q}_l^+\approx\mathbf{I}$, pre-synaptic inputs may also be estimated approximately such that
\begin{equation}
  \hat{\mathbf{a}}_{l-1}:= g_l^{-1}\left(\mathbf{z}_l-g_l(\mathbf{y}\mathbf{U}_l)\right)\mathbf{Q}_l^{+}.
\end{equation}
Thus, pre-synaptic inputs are encoded in a lossy manner in each layer's neural pre-activation potentials. In our experiments, we employed the sign function for target generation, defined as:
\begin{equation}
g_l:=\operatorname{sign}(x)=\begin{cases}
-1 & x<0\\
0 & x=0\\
1&x>0
\end{cases}
\end{equation}
The element-wise function $g_l:=\operatorname{sign}(x)$ was selected due to computational simplicity and predictable distribution. The $\operatorname{tanh}$ function was employed as a heuristic surrogate for $g_l^{-1}$, yielding satisfactory results.

\subsection{Stability Experiments}\label{remark: Stability Analysis}
\begin{figure}
  \centering
  \includegraphics[width=0.9\linewidth]{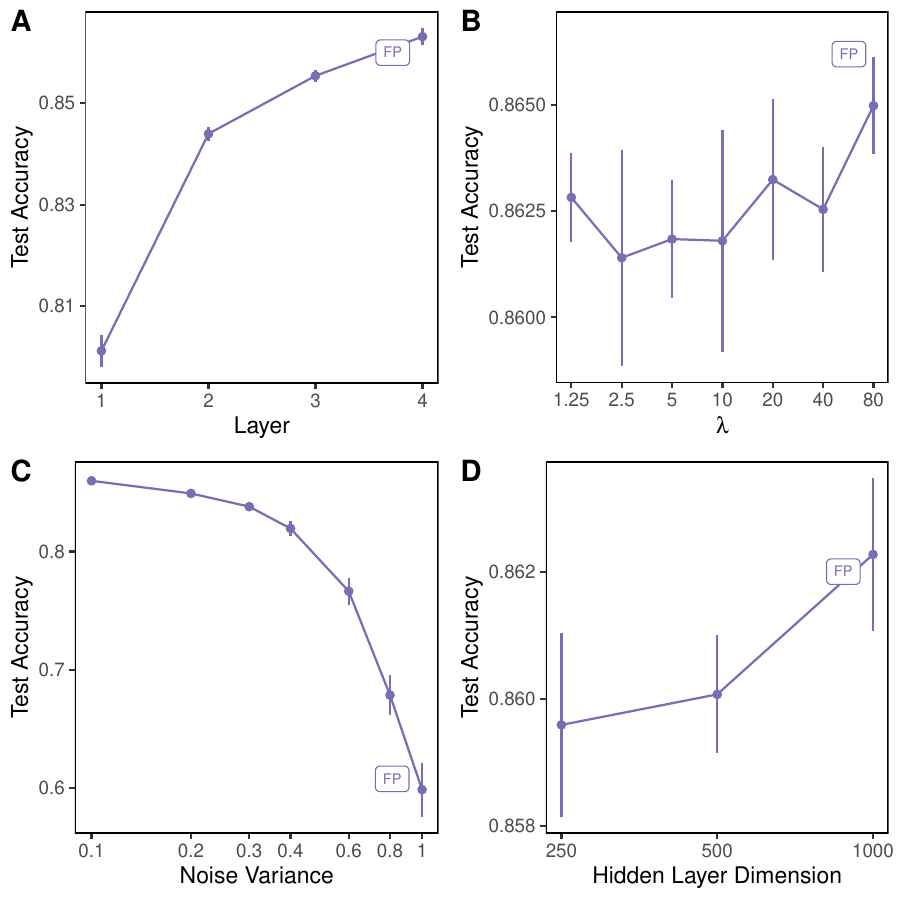}
   \caption{Technical analysis of multilayer perceptron trained by Forward Projection. A: Test accuracy of hidden layer explanations in a $4\times 1000$ neuron MLP trained by FP on FMNIST.  B: Test accuracy of $3\times 1000$ neuron MLP trained trained by FP on FMNIST using different $\lambda$ regularisation values. C: Test accuracy of $3\times 1000$ neuron MLP trained trained by FP on FMNIST with additive Gaussian noise in training data. D: Variability of FP training performance using different random projections in  $3\times 1000$ neuron MLP on FMNIST. Points represent mean over all replicates and error bars represent $\pm$ one standard deviation. 
   BP: backpropagation; DTP: Difference Target Propagation; FF: Forward-Forward; FP: Forward Projection; LP: label projection; LPN: noisy label projection; LS: Local Supervision; PC: Predicitve Coding; RF: Random Features.}
  \label{fig: mlp technical analysis}
\end{figure}

Technical analysis and stability experiments were performed on MLP models training on the FMNIST data. The stability of layer explanations, on models fitted to different data folds, with different random projection matrices, using a $4\times 1000$ neuron feedforward architecture was evaluated. Test performance of layer explanations improved progressively from the first layer to the final layer with low variability throughout (Figure \ref{fig: mlp technical analysis}-A). Model stability with respect to choices of the FP penalty parameter $\lambda$ (Equation \ref{eq:fp fitting}) in hidden layers was evaluated in a $3\times 1000$ neuron MLP. Test accuracies were stable from $\lambda=1.25$ to $\lambda=80$  with the highest observed performance at $\lambda=80$ (Figure \ref{fig: mlp technical analysis}-B). The mean variance of pixels in the FMNIST training dataset is $\sigma_{\operatorname{input}}=0.087$. Although additive gaussian noise caused progressive deterioration in model performance from $\sigma_{\operatorname{noise}}=0.1$ to $\sigma_{\operatorname{noise}}=1$,  test accuracy remained above $0.819$ for all  $\sigma_{\operatorname{noise}}<0.4$, which corresponds to a signal-to-noise ratio of approximately $1:4.6$ (Figure \ref{fig: mlp technical analysis}-C). The impact of the randomised projections on model performance was assessed in 3-layer MLP models with $2\times1000$ hidden neurons in the first two layers and $m_3\in\{250, 500, 1000\}$ in the penultimate layer, with stable test performance observed throughout (Figure \ref{fig: mlp technical analysis}-D).

\subsection{Optimisation Theory}\label{remark:opt theory}
We now compare the ordinary least squares linear model to a two-layer network trained by Forward Projection and a Random Features model. We aim to demonstrate analytically that the expected error of the Forward Projection model improves upon the others.

Consider a dataset $\{\mathbf{x}_i, \mathbf{y}_i\}_{i=1}^N$, where $\mathbf{x}_i \sim \mathcal{N}(\mathbf{0}, \mathbf{I}_{m_l})$ and target $\mathbf{y}_i\in \mathbb R ^{m_L}$. Let $\mathbf{X} \in \mathbb R ^{N\times m_l}$ denote the design matrix with rows $\mathbf{x}_i^\top$ and $\mathbf{Y} \in \mathbb R ^{N\times m_L}$ denote the target matrix with rows $\mathbf{y}_i^\top$. Let $\operatorname{col}(\mathbf{X})$ denote the column space of matrix $\mathbf{X}$, the span of its column vectors. We assume $N>m_l\gg m_L$  and that $\mathbf{X}$ has full column rank. We first consider the expected error of a single-layer, linear model, with weights $\hat{\mathbf{W}}_{\mathrm{linear}}\in \mathbb R^{m_l\times m_L}$. We estimate $\hat{\mathbf{W}}_{\mathrm{linear}}$, by ordinary least squares \cite{Anderson2003-lc}, such that
\begin{equation}
    \hat{\mathbf{W}}_{\mathrm{linear}}:=(\mathbf{X}^\top\mathbf{X})^{-1}\mathbf{X}^\top\mathbf{Y}.
\end{equation}
We denote the projection matrix of $\mathbf{X}$ as 
\begin{equation}
    \mathbf{P}_\mathbf{X}=\mathbf{X}(\mathbf{X}^\top\mathbf{X})^{-1}\mathbf{X}^\top.
\end{equation}
The spectrum of the idempotent projection matrix $\mathbf{P}_\mathbf{X}$ is composed of a unit eigenvalue with multiplicity $m_l$ and a zero eigenvalue with multiplicity $N-m_l$ \cite{Meyer2000-yo}.  Letting $\hat{\mathbf{Y}}_{\mathrm{linear}}=\mathbf{P}_\mathbf{X}\mathbf{Y}$, the model error is given by
\begin{equation}
E_{\mathrm{linear}}=\mathbb E\left[ \lVert\mathbf{Y}-\hat{\mathbf{Y}}_{\mathrm{linear}}\rVert^2_F\right]=\mathbb E\left[ \lVert(\mathbf{I}-\mathbf{P}_{\mathbf{X}})\mathbf{Y}\rVert^2_F\right],
\end{equation}
\textcolor{black}{Letting $\mathbf{Y}_\perp=(\mathbf{I}-\mathbf{P}_{\mathbf{X}})\mathbf{Y}$, we decompose \cite{Horn2012} 
\begin{equation}
    \mathbf{Y}=\mathbf{P}_{\mathbf{X}}\mathbf{Y}+(\mathbf{I}-\mathbf{P}_{\mathbf{X}})\mathbf{Y}=\hat{\mathbf{Y}}_{\mathrm{linear}}+\mathbf{Y}_\perp\label{eq y decomp},
\end{equation}
where $\operatorname{col}(\hat{\mathbf{Y}}_{\mathrm{linear}})\subset \operatorname{col}(\mathbf{X})$, and $\mathbf{Y}_\perp$ contains residuals. Let $\operatorname{col}(\mathbf{X})^\perp$ denote the $(N-m_l)$-dimensional space orthogonal to $\operatorname{col}(\mathbf{X})$. We note that  as $(\mathbf{I}-\mathbf{P}_{\mathbf{X}})$ projects onto $\operatorname{col}(\mathbf{X})^\perp$, $\mathbf{Y}_\perp$ lies in $\operatorname{col}(\mathbf{X})^\perp$. By symmetric idempotency of $(\mathbf{I}-\mathbf{P}_{\mathbf{X}})$ and cyclic trace invariance \cite{Horn2012}, we have \begin{multline}
E_{\mathrm{Linear}}=\mathbb{E}\big[\lVert\mathbf{Y}_\perp\rVert_F^2\big]
    = \mathbb{E}\left[\operatorname{Tr}\left((\mathbf{I}-\mathbf{P}_{\mathbf{X}})\,\mathbf{Y}\mathbf{Y}^\top(\mathbf{I}-\mathbf{P}_{\mathbf{X}})^\top\right)\right]
\\ = \mathbb{E}\left[\operatorname{Tr}\left((\mathbf{I}-\mathbf{P}_{\mathbf{X}})^\top(\mathbf{I}-\mathbf{P}_{\mathbf{X}})\,\mathbf{Y}\mathbf{Y}^\top\right)\right]
= \mathbb{E}\left[\operatorname{Tr}\left((\mathbf{I}-\mathbf{P}_{\mathbf{X}})\,\mathbf{Y}\mathbf{Y}^\top\right)\right].\label{eq E linear}
\end{multline}
We now consider a two-layer neural network model trained by forward projection for fitting the same data. The model has $m_l$ hidden $\operatorname{ReLU}$-activated neurons, such that
\begin{equation}
\hat{\mathbf{Y}}_{\mathrm{FP}}:=\operatorname{ReLU}(\mathbf{X}\hat{\mathbf{W}}_1)\hat{\mathbf{W}}_2,\label{eq ydef}
\end{equation}
The task is to identify weight matrices $\hat{\mathbf{W}}_1\in \mathbb R^{m_l\times m_l}, \hat{\mathbf{W}}_2\in \mathbb R^{m_l\times m_L}$ to minimise the error $
\lVert \mathbf{Y}-\hat{\mathbf{Y}}_{\mathrm{FP}}\rVert^2_F
$. Let $\mathbf{U}\in\mathbb R^{m_L\times m_l}$ and  $\mathbf{Q}\in \mathbb R^{m_l\times m_l}$ be matrices with elements sampled from $\mathcal N(0,1)$. We now fit $\hat{\mathbf{W}}_1$ to estimate $\mathbf{XQ}+\mathbf{YU}$ from $\mathbf{X}$, such that
\begin{equation}
    \hat{\mathbf{W}}_1:=(\mathbf{X}^\top\mathbf{X})^{-1}\mathbf{X}^\top(\mathbf{XQ}+\mathbf{YU}).\label{eq: ot2}
\end{equation}
Thus, the pre-activation potential at the first layer is given by
\begin{equation}
\mathbf{X}\hat{\mathbf{W}}_1=\mathbf{XQ} +\mathbf{P}_\mathbf{X}\mathbf{YU}.\label{eq: ot4}
\end{equation}
 As $m_l\gg m_L$, we assume $\textbf{UU}^+=\textbf{I}_{m_L}$.  Independence of $ \mathbf{Q}$ and $\mathbf{U}$, with elements sampled from $\mathcal{N}(0,1)$ gives $\mathbb E_{\mathbf{Q}, \mathbf{U}}\left[\mathbf{QU}^+\right]=0$. Therefore,
\begin{equation}
    \mathbb E_{\mathbf{Q}, \mathbf{U}}\left[\mathbf{X}\hat{\mathbf{W}}_1\mathbf{U}^+\right]=
    \mathbb E_{\mathbf{Q}, \mathbf{U}}\left[\mathbf{XQU}^+\right]+\mathbb E_{ \mathbf{U}}\left[\mathbf{P}_\mathbf{X}\mathbf{YUU}^+\right]=\mathbf{P}_\mathbf{X}\mathbf{Y}.
\end{equation}
Therefore, we note that in the $m_l\gg m_L$ setting, $\mathbf{P}_{\mathbf{X}}\mathbf{Y}$ is encoded by random projection over $\mathbf{X}\hat{\mathbf{W}}_1$.
}

\textcolor{black}{
Let $\mathbf{A}=\operatorname{ReLU}(\mathbf{X}\hat{\mathbf{W}}_1)$ denote the hidden activations. We note that $\operatorname{col}(\mathbf{A}) \ne\operatorname{col}(\mathbf{X})$, as the $\operatorname{ReLU}$ operation introduces non-linearities into $\mathbf{X}\hat{\mathbf{W}}_1$. Since $(\mathbf{Q}, \mathbf{U})$ is sampled from a symmetric distribution around zero, the realisations $(\mathbf{Q}, \mathbf{U})$ and $(\mathbf{-Q}, \mathbf{-U})$ have equal probability. Using the identity
\begin{equation}
\mathbf{X}\hat{\mathbf{W}}_1=\operatorname{ReLU}(\mathbf{X}\hat{\mathbf{W}}_1)-\operatorname{ReLU}(-\mathbf{X}\hat{\mathbf{W}}_1),    
\end{equation}we have 
\begin{equation}
    \mathbb E_{\mathbf{Q}, \mathbf{U}}\left[\operatorname{ReLU}(\mathbf{X}\hat{\mathbf{W}}_1)\mathbf{U}^+ \right]=  -\mathbb E_{\mathbf{Q}, \mathbf{U}}\left[\operatorname{ReLU}(-\mathbf{X}\hat{\mathbf{W}}_1)\mathbf{U}^+ \right].
\end{equation}
Therefore 
\begin{equation}
     \mathbb E\left[\mathbf{A}\mathbf{U}^+\right]= \frac{1}{2}\mathbb E\left[\mathbf{X}\hat{\mathbf{W}}_1\mathbf{U}^+\right]=\frac{1}{2}\mathbf{P}_\mathbf{X}\mathbf{Y}.
\end{equation}
In the $m_l\gg m_L$ setting, we assume convergence to the mean, and approximate $\mathbf{A}\mathbf{U}^+\approx\frac{1}{2}\mathbf{P}_\mathbf{X}\mathbf{Y}$. Accordingly, we assume that $\mathbf{P}_\mathbf{X}\mathbf{Y}$ lies within $\operatorname{col}(\textbf{A})$  -- i.e.,  $\operatorname{col}(\hat{\mathbf{Y}}_{\mathrm{linear}})\subset\operatorname{col}(\mathbf{A})$. We also assume $\mathbf{A}$ has full column rank. We now fit $\hat{\mathbf{W}}_2$ to estimate $\mathbf{Y}$ from $\mathbf{A}$, such that
\begin{equation}
    \hat{\mathbf{W}}_2:=(\mathbf{A}^\top\mathbf{A})^{-1}\mathbf{A}^\top\mathbf{Y}.\label{eq: ot2 fp}
\end{equation}
Letting $\hat{\mathbf{Y}}_{\mathrm{FP}}=\mathbf{P}_\mathbf{A}\mathbf{Y}$, the Forward Projection error is given by
\begin{equation}
E_{\mathrm{FP}}=\mathbb E\left[ \lVert\mathbf{Y}-\hat{\mathbf{Y}}_{\mathrm{FP}}\rVert^2_F\right]=\mathbb E\left[ \lVert(\mathbf{I}-\mathbf{P}_{\mathbf{A}})\mathbf{Y}\rVert^2_F\right].\label{eq e fp}
\end{equation}
 From \eqref{eq y decomp}, we have
\begin{equation}
    (\mathbf{I}-\mathbf{P}_{\mathbf{A}})\mathbf{Y}=(\mathbf{I}-\mathbf{P}_{\mathbf{A}})\hat{\mathbf{Y}}_{\mathrm{linear}}+(\mathbf{I}-\mathbf{P}_{\mathbf{A}})\mathbf{Y}_\perp. \label{eq y decomp2}
\end{equation}
From our assumption that $\operatorname{col}(\hat{\mathbf{Y}}_{\mathrm{linear}})\subset\operatorname{col}(\mathbf{A})$, we have
\begin{equation}
(\mathbf{I}-\mathbf{P}_{\mathbf{A}})\hat{\mathbf{Y}}_{\mathrm{linear}}= 0.    \end{equation}
Substituting \eqref{eq e fp} and \eqref{eq y decomp2}, we have 
\begin{equation}
   E_{\mathrm{FP}}=\mathbb E\left[ \lVert(\mathbf{I}-\mathbf{P}_{\mathbf{A}})\mathbf{Y}_\perp\rVert^2_F\right].
\end{equation}
As $\mathbf{P}_{\mathbf{A}}\mathbf{Y}_\perp$ is orthogonal to $(\mathbf{I}-\mathbf{P}_{\mathbf{A}})\mathbf{Y}_\perp$, we may apply the Pythagorean theorem \cite{Kreyszig1989-xv}, such that
\begin{equation}
    \lVert \mathbf{Y}_\perp\rVert^2_F=\lVert \mathbf{P}_{\mathbf{A}}\mathbf{Y}_\perp\rVert^2_F+\lVert (\mathbf{I}-\mathbf{P}_{\mathbf{A}})\mathbf{Y}_\perp\rVert^2_F.
\end{equation}
Substituting \eqref{eq E linear} and \eqref{eq e fp}, we have
\begin{equation}
   E_{\mathrm{linear}}= \mathbb E\left[\lVert \mathbf{P}_{\mathbf{A}}\mathbf{Y}_\perp\rVert^2_F\right]+E_{\mathrm{FP}}.
\end{equation}
Therefore, under the inclusion assumption that $\operatorname{col}(\hat{\mathbf{Y}}_{\mathrm{linear}})\subset \operatorname{col}(\mathbf{A})$, the improvement of the Forward Projection model over the linear model is given by
\begin{equation}
\boxed{
   E_{\mathrm{linear}}-E_{\mathrm{FP}}= \mathbb E\left[\lVert \mathbf{P}_{\mathbf{A}}\mathbf{Y}_\perp\rVert^2_F\right]\ge 0. }\label{eq linfp}
\end{equation}
}

\textcolor{black}{
For comparison, we also consider the performance of a two-layer neural network model trained with random weights in the first layer (i.e., Random Features), such that $\mathbf{W}_{\mathrm{random}}\in\mathbb R^{m_l\times m_l}$ has elements sampled from $\mathcal{N}(0,1)$. The hidden activities are given by \begin{equation}
\mathbf{A}_{\mathrm{random}}=\operatorname{ReLU}\left(\mathbf{X}\mathbf{W}_{\mathrm{random}}\right).    
\end{equation}
Since $\mathbf{W}_{\mathrm{random}}$ is independent of $\mathbf{Y}$, the above hidden representation lacks alignment with the task-relevant structure that Forward Projection encodes due to the $\mathbf{YU}$ term. Hence, we cannot assume $\operatorname{col}(\hat{\mathbf{Y}}_{\mathrm{linear}})\subset \operatorname{col}(\mathbf{A}_{\mathrm{random}})$. Lastly, due to the random selection of $\mathbf{W}_{\mathrm{random}}$, we have 
\begin{equation}
\mathbb P\left[(\mathbf{A}_{\mathrm{random}})_{ij}=0\right]=\mathbb P\left[(\mathbf{X}\mathbf{W}_{\mathrm{random}})_{ij}\le 0 \right]=\frac{1}{2}    
\end{equation}
Therefore, $\mathbf{A}_{\mathrm{random}}$ represents a random, nonlinear transformation of $\mathbf{X}$, with equal dimension, and reduced representational capacity due to element-wise sparsity. Only the output layer $\textbf{W}_{2}$ may be trained, such that:
\begin{equation}
    \hat{\mathbf{W}}_2:=(\mathbf{A}_{\mathrm{random}}^\top\mathbf{A}_{\mathrm{random}})^{-1}\mathbf{A}_{\mathrm{random}}^\top\mathbf{Y}.
\end{equation}
Unlike $\mathbf{A}$, which is correlated with the label, there is no reason to expect, over the distribution of $\mathbf{W}_{\mathrm{random}}$,  that $\mathbf{A}_{\mathrm{random}}$ is a more informative design matrix than $\mathbf{X}$. Therefore, the error of this Random Features network is not expected to improve upon $E_{\mathrm{linear}}$. We note that in the special case examined here, the hidden dimension is equal to the input dimension, a framework which contrasts with the typical usage of Random Features  \cite{Rahimi2007}, where high-dimensional random projection is employed to generate a large, representative feature set. Unlike Forward Projection, which provides the improvement in  \eqref{eq linfp}, Random Features hidden layer activations do not provide such an expectation in a hidden layer of equal dimension to the input. Hence, this comparison reveals that ReLU nonlinearity alone is insufficient for expected improvement -- the label-informed initialization is also essential. 
}

\textcolor{black}{
This analysis compares a regression model with $m_l\cdot m_L$ parameters in a single layer trained by ordinary least squares regression to a two-layer neural network with $m_l(m_l+m_L)$ parameters, trained by Forward Projection; and a two-layer neural network with $m_l$ random features and $m_lm_L$ trainable parameters. Although the Forward Projection model has the greatest number of free parameters, it is noted that these are not jointly optimised, but instead may be separated into $m_l$ linear regressions in $\mathbf{W}_1$ and, subsequently, $m_L$ linear regressions in $\mathbf{W}_2$.
}

\textcolor{black}{
Differences in the modelling of Forward Projection, Random Features and linear regression each present distinct profiles of advantages and limitations. Despite these architectural differences, this analysis establishes that Forward Projection offers a principled, backpropagation-free training method with expected improvement over linear regression -- an expectation absent from Random Features of equal dimension.
}

\subsection{Iterative learning with Forward Projection}\label{remark: iterative fp} 
An alternative to deriving the closed-form Forward Projection solution for regression over the entire dataset is to approximate the optimization procedure through iterative updates on mini-batches, while preserving local learning dynamics and target generation as formalized in Equation \ref{eq:fp target gen}. We partition the dataset $\mathcal{D}$ into $T$ mini-batches $\mathcal{B}_t = \{(\mathbf{x}_j, \mathbf{y}_j)\}_{j=1}^{B}$, where $B$ is the batch size. For each mini-batch, compute $
\mathbf{A}_{l-1}^{(t)} \in \mathbb{R}^{B \times m_{l-1}}$ and $\tilde{\mathbf{Z}}_l^{(t)} \in \mathbb{R}^{B \times m_l}$ using the target generation rule
\begin{equation}
\tilde{\mathbf{z}}_l = g_l(\mathbf{a}_{l-1}\mathbf{Q}_l) + g_l(\mathbf{y}\mathbf{U}_l).
\end{equation}
For layer $l$, define the local loss:
\begin{equation}
\mathcal{L}_l^{(t)} = \frac{1}{B}\sum_{j=1}^{B} \|\mathbf{z}_{j,l} - \tilde{\mathbf{z}}_{j,l}\|^2,   
\end{equation}
where $\mathbf{z}_{j,l} = \mathbf{a}_{j,l-1}\mathbf{W}_l$. The gradient of $\mathcal{L}_l^{(t)}$ with respect to $\mathbf{W}_l$ is \cite{Golub2013-lw}:
\begin{equation}
\nabla_{\mathbf{W}_l} \mathcal{L}_l^{(t)} = \frac{2}{B} (\mathbf{A}_{l-1}^{(t)})^\top (\mathbf{A}_{l-1}^{(t)} \mathbf{W}_l - \tilde{\mathbf{Z}}_l^{(t)}).
\end{equation}
Therefore, weights may be updated according to 
\begin{equation}
\mathbf{W}_l \leftarrow \mathbf{W}_l - \eta \nabla_{\mathbf{W}_l} \mathcal{L}_l^{(t)},
\end{equation}
where $\eta$ is the learning rate. Optionally, regularisation may be included via the ridge penalty \cite{Golub2013-lw}:
\begin{equation}
\nabla_{\mathbf{W}_l} \mathcal{L}_l^{(t)} = \frac{2}{B} \left(\mathbf{A}_{l-1}^{(t)}\right)^\top \left(\mathbf{A}_{l-1}^{(t)} \mathbf{W}_l - \tilde{\mathbf{Z}}_l^{(t)}\right) + \frac{2}{B}\lambda \mathbf{W}_l.
\end{equation}
In experiments on FMNIST classification by MLPs with $3\times1000$ hidden neurons, sequential FP training achieved similar test performance to the closed-form FP solution (test AUC: $98.6 \pm 0.79\%$, test accuracy: $84.4\pm 0.02\%$).

\subsection{Modelling high-dimensional labels}\label{remark: modelling hd} 
We now analyse the dependence of the expected local error on the layer and label dimensions. From \eqref{eq: fp label estimation}, we have,
\begin{equation}
  \hat{\mathbf{Y}}_l := g_l^{-1}\left(\mathbf{Z}_l-g_l(\mathbf{A}_{l-1}\mathbf{Q}_l)\right)\mathbf{U}_l^+.
\end{equation}
For simplicity, we analyse the special case that $g_l$ is the identity function yielding,
\begin{equation}
  \hat{\mathbf{Y}}_l :=\left(\mathbf{Z}_l-\mathbf{A}_{l-1}\mathbf{Q}_l\right)\mathbf{U}_l^+.
\end{equation}
Recalling that the realised potentials are estimates of pre-defined target potentials, such that $\mathbf{Z}_l=\tilde{\mathbf{Z}}_l-\mathbf{E}_l$, we have,
\begin{equation}
  \hat{\mathbf{Y}}_l :=\left(\tilde{\mathbf{Z}}_l-\mathbf{E}_l-\mathbf{A}_{l-1}\mathbf{Q}_l\right)\mathbf{U}_l^+.
\end{equation}
Applying the identity function as $g_l$ in equation  \eqref{eq:fp target gen},  we have,
\begin{equation}
 \tilde{\mathbf{Z}}_l=\mathbf{A}_{l-1}\mathbf{Q}_l+\mathbf{Y}\mathbf{U}_l.
 \end{equation}
 Substituting, we have,
\begin{equation}
  \hat{\mathbf{Y}}_l =\left(\mathbf{Y}\mathbf{U}_l-\mathbf{E}_l\right)\mathbf{U}_l^+.
\end{equation}
Assuming that $\mathbf{U}$ has full row rank, and that $m_l\ge m_L$, we have $\mathbf{U}_l\mathbf{U}^+_l=\mathbf{I}_{m_L}$ \cite{Penrose1955}. Therefore, \begin{equation}
  \hat{\mathbf{Y}}_l =\mathbf{Y}-\mathbf{E}_l\mathbf{U}_l^+.\label{eq y minus eu}
\end{equation}
\textcolor{black}{As the realised potential is given by $\mathbf{Z}_l=\mathbf{P}_{\mathbf{A}_{l-1}}\tilde{\mathbf{Z}}_l$, the error is given by \begin{equation}
    \mathbf{E}_l=\tilde{\mathbf{Z}}_l-\mathbf{Z}_l=\left(\mathbf{I}-\mathbf{P}_{\mathbf{A}_{l-1}}\right)\tilde{\mathbf{Z}}_l=\left(\mathbf{I}-\mathbf{P}_{\mathbf{A}_{l-1}}\right)\left(\mathbf{A}_{l-1}\mathbf{Q}_{l} + \mathbf{Y}\mathbf{U}_l\right),
\end{equation}where $\mathbf{P}_{\mathbf{A}_{l-1}}=\mathbf{A}_{l-1}\left(\mathbf{A}_{l-1}^\top\mathbf{A}_{l-1} \right)^{-1}\mathbf{A}_{l-1}^\top$ is the projection matrix of the previous layer activity $\mathbf{A}_{l-1}$. As $\left(\mathbf{I}-\mathbf{P}_{\mathbf{A}_{l-1}}\right)\mathbf{A}_{l-1}=0$, we have
\begin{equation}
     \mathbf{E}_l=\left(\mathbf{I}-\mathbf{P}_{\mathbf{A}_{l-1}}\right)\mathbf{Y}\mathbf{U}_l.
\end{equation}
The squared error is given by the squared Frobenius norm
\begin{equation}
      \lVert \mathbf{E}_l \rVert^2_F=\operatorname{Tr}\left(\mathbf{U}_l^\top\mathbf{Y}^\top\left(\mathbf{I}-\mathbf{P}_{\mathbf{A}_{l-1}}\right)\mathbf{Y}\mathbf{U}_l\right).
\end{equation}
As $\mathbf{U}_l\in \mathbb R ^{m_L\times m_l}$ has entries sampled from $\mathcal{N}(0,1)$, we have the expectation
\begin{equation}
     \mathbb E_{\mathbf{U}_l}\left[ \lVert \mathbf{E}_l \rVert^2_F \right]=m_l\cdot \lVert\left(\mathbf{I}-\mathbf{P}_{\mathbf{A}_{l-1}}\right)\mathbf{Y}\rVert^2_F.
\end{equation}
Firstly, it is observed that $ \mathbb E\left[ \lVert \mathbf{E}_l \rVert^2_F \right]$ grows linearly with the number of columns in $\mathbf{Y}$, i.e. $m_L$. Secondly, it is observed that  $ \mathbb E\left[ \lVert \mathbf{E}_l \rVert^2_F \right]$ grows linearly with the number of columns in $m_l$, which corresponds to equal expected error in each column of $\textbf{Z}_l$. Assuming $\mathbf{U}_l\mathbf{U}_l^+=\mathbf{I}_{m_L}$, we have \begin{equation}
  \mathbf{E}_l\mathbf{U}_l^+= \left(\mathbf{I}-\mathbf{P}_{\mathbf{A}_{l-1}}\right)\mathbf{Y}\mathbf{U}_l\mathbf{U}_l^+=\left(\mathbf{I}-\mathbf{P}_{\mathbf{A}_{l-1}}\right)\mathbf{Y}
\end{equation}
Substituting into \eqref{eq y minus eu}, and taking the squared frobenius norm we have the squared error of the label prediction
\begin{equation}
      \mathbb E_{\mathbf{Q}_l, \mathbf{U}_l}\left[\lVert\mathbf{Y}-\hat{\mathbf{Y}}_l \rVert^2_F\right]=
      \mathbb E_{\mathbf{Q}_l, \mathbf{U}_l}\left[\lVert\mathbf{E}_l\mathbf{U}_l^+\rVert^2_F\right]=\lVert\left(\mathbf{I}-\mathbf{P}_{\mathbf{A}_{l-1}}\right)\mathbf{Y}\rVert^2_F
\end{equation}
Therefore $\mathbb E_{\mathbf{Q}_l, \mathbf{U}_l}\left[\lVert\mathbf{Y}-\hat{\mathbf{Y}}_l \rVert^2_F\right]$ increases with the component of $\mathbf{Y}$ orthogonal to $\operatorname{col}(\mathbf{A}_{l-1})$. As $\operatorname{rank}(\mathbf{A}_{l-1})\le m_{l-1}$, enlarging $m_{l-1}$ may increase the capacity for $\operatorname{col}(\mathbf{A}_{l-1})$ to represent $\mathbf{Y}$. 
}  

\subsection{Training complexity of other methods} \label{remark: train complexity other}
In this section, we describe the training complexity involved in each step of the different methods that are used in this paper for benchmarking purposes. Hereafter, $N$ refers to the sample size, $N_e$ to the number of training epochs, $m$ the hidden layer dimension (it is assumed all layers have dimension $m$), $m_L$ to label dimension and $B$ to batch size.

\subsubsection{Backpropagation}
Backpropagation requires a forward pass through each layer ($N_eNm^2$), a backward pass to compute weight gradient ($N_eNm^2$) and the input gradient ($N_eNm^2$). After each batch, weights are updated ($N_eB^{-1}Nm^2$). For each layer, backpropation requires storage of weight parameters ($m^2$), input activations ($Bm$), and weight gradients ($m^2$).

\subsubsection{Local Supervision}
In local supervision, after the forward pass through the main layer ($N_eNm^2$ MACs), an auxiliary forward pass is conducted through the local supervision head  ($N_eNmm_L$ MACs). Subsequently, the auxiliary gradient is computed at the local supervision head ($N_eNmm_L$ MACs), which is backpropagated through the main layer ($N_eNm^2$ MACs).  Forward weight updates for the main layer occur at the end of each batch ($N_eB^{-1}Nm^2$ MACs). Optionally, the auxiliary layer may be trained, requiring gradient computation ($N_eNmm_L$ MACs) and weight updates ($N_eB^{-1}Nmm_L$ MACs). Local supervision requires storage of main layer weights ($m^2$), auxiliary layer weights ($mm_L$), main layer input activations ($Bm$), main layer output activations ($Bm$), auxiliary layer output activations ($Bm_L$) and weight updates ($m^2$).

\subsubsection{Forward-Forward (FF)}
For a given layer, FF requires two forward passes -- for ``positive" and ``negative" data, respectively ($2N_eNm^2$ MACs) \cite{Hinton2022Dec}. A goodness metric is computed for both positive and negative activations, requiring $2N_eNm$ MACs. Local gradients are computed for both positive and negative passes ($2N_eNm^2$ MACs). Forward weight updates occur with each batch ($N_eB^{-1}Nm^2$ MACs). At inference time, forward passes are required for each possible class label ($Nm^2m_L$ MACs). Forward-Forward requires storage of weight parameters ($m^2$), positive and negative input activations ($2Bm$), positive and negative output activations ($2Bm$) and weight updates ($m^2$). 

\subsubsection{Difference Target Propagation (DTP)}
DTP replaces gradient transport with targets formed using a learned approximate inverse (decoder) and a difference correction. Training requires an encoder forward pass ($N_eNm^2$ MACs), and two decoder passes ($2N_eNm^2$ MACs) to perform the difference correction operation between targets and observed activations \cite{Lee2014}. Decoder training employs a denoising style loss, forwarding through both encoder ($N_eNm^2$ MACs) and decoder ($N_eNm^2$ MACs), before computing the decoder weight gradient ($N_eNm^2$ MACs). Finally, the encoder weight gradient is computed with respect to the local target loss ($N_eNm^2$ MACs). Weight updates for both encoder and decoder occur after each batch ($2N_eB^{-1}Nm^2$ MACs). At inference time, only the encoder forward pass is required. Assuming batch size $B=1$, training requires a forward pass ($N_eNm^2$ MACs) and $8N_eNm^2$ further MACs per layer. DTP requires storage of encoder and decoder weight parameters ($2m^2$), input activations ($Bm$), output activations ($Bm$), targets ($Bm$) and weight updates ($m^2$), which can be performed sequentially for encoder and decoder.

\subsubsection{Predictive Coding}
PC performs \(k\) iterations of activity inference that alternate forward prediction ($kN_eNm^2$ MACs) and feedback error projection through a decoder network ($kN_eNm^2$ MACs), followed by a local Hebbian weight gradient  computation ($N_eNm^2$ MACs) \cite{Millidge2021Jul}. Forward weight updates occur at the end of each batch ($N_eB^{-1}Nm^2$ MACs). Letting $k=1$, $B=1$, training requires a forward pass ($N_eNm^2$ MACs) and $3N_eNm^2$ further MACs per layer.  Predictive coding requires storage of weight parameters ($m^2$), input activations ($Bm$), output activations ($Bm$),  output targets ($Bm$) and hebbian weight updates ($m^2$).

\subsection{Modelling with Challenging Hidden Activation Functions}\label{challenging activations}
Neural networks commonly employ simple activation functions such as $\operatorname{ReLU}$, due to low computational complexity and favourable approximation properties. Alternative hidden activation functions in sigmoid, polynomial and modulo families are desirable for modelling various physical and theoretical systems \cite{Wang2022Feb}. However, problems such as gradient vanishing and saturation may arise when these activation functions are trained via SGD-based methods \cite{Glorot2010Mar, Wang2022Feb}. SGD-based training is not directly applicable where $f_l$ is undifferentiable, necessitating the use of surrogate gradient methods. Many undifferentiable activation functions, such as the Heaviside step function, have attractive properties of low computational complexity and biological plausibility. Forward Projection does not require hidden activation functions to be differentiable, as target potentials $\tilde{\mathbf{Z}}_1,\ldots, \tilde{\mathbf{Z}}_{l-1}$ are modelled before activation. Thus, Forward Projection presents many opportunities for modelling activation functions for which SGD-based training is intractable. Going beyond the standard activation functions, we evaluated the performance of functions which present a challenge for SGD-based training.
Networks were modelled with modulo 2 activation (``mod2": $f(x)= x \mod 2$) and ``square" activation $f(x)=x^2$. To control the desired range for the target potentials, an additive constant $\alpha\in \mathbb R$ was included in the target generation function, such that
\begin{equation}
 \tilde{\mathbf{z}}_l=\operatorname{sign}(\mathbf{a}_{l-1}\mathbf{Q}_l)+\operatorname{sign}(\mathbf{yU}_l)+\alpha.\label{eq:fp_target_gen2}
\end{equation}
This hyperparameter was predefined as $\alpha=0.5$ for both mod2 and square implementations. Forward projection yielded consistent performance across various activation functions in each dataset. In contrast, attempts at SGD-based training failed to converge in each task. Performance of Forward Projection, random features, Local Supervision, Forward-Forward and backpropagation modelling with square and mod2 activations is provided in Supplementary Table \ref{tab:challenging_activations}.

\begin{table}
\begin{tabular}{p{1.3cm}p{1.2cm}p{1.6cm}p{1.3cm}p{1.5cm}p{1.5cm}p{1.5cm}}
\toprule
 & & \multicolumn{4}{r}{} \\
 & Method & Forward projection (Ours) &Random features&Local Supervision&Forward Forward& Backprop. (reference standard) \\
Dataset & Activation & & & & \\
\midrule
\multirow[t]{2}{*}{FMNIST} & mod2 & 60.9 ± 0.4 & 34.4 ± 0.4 & 10.0 ± 0.2 & 9.8 ± 0.3 & 9.9 ± 0.2 \\
 & square & 86.0 ± 0.3 & 67.0 ± 0.2 & 81.5 ± 1.4 & 38.6 ± 39.1 & 65.9 ± 6.2 \\
\multirow[t]{2}{*}{Promoters} & mod2 & 76.7 ± 0.4 & 49.7 ± 0.6 & 50.0 ± 0.0 & 50.2 ± 0.5 & 49.9 ± 0.2 \\
 & square & 78.4 ± 6.2 & 50.0 ± 0.0 & 52.2 ± 3.0 & 54.4 ± 9.8 & 54.5 ± 5.7 \\
\multirow[t]{2}{*}{PTBXL-MI} & mod2 & 76.6 ± 2.5 & 50.0 ± 0.3 & 50.0 ± 0.0 & 48.5 ± 2.2 & 50.0 ± 0.0 \\
 & square & 83.3 ± 1.5 & 50.0 ± 0.0 & 51.0 ± 2.0 & 64.6 ± 4.7 & 51.4 ± 1.6 \\
\bottomrule
\end{tabular}

\caption{Test accuracies of local learning methods in FashionMNIST, Promoters and PTBXL-MI tasks, using modulo 2 (``mod2") ($f(x)=x\mod{2}$) and square ($f(x)=x^{2}$) activation functions. FP: Forward Projection; LS: Local Supervision; FF: Forward-Forward; BP: Backpropagation}
\label{tab:challenging_activations}

\end{table}

\subsection{Few-shot modelling}\label{remark:fewshot}
\begin{figure}
  \centering
  \includegraphics[width=1\linewidth]{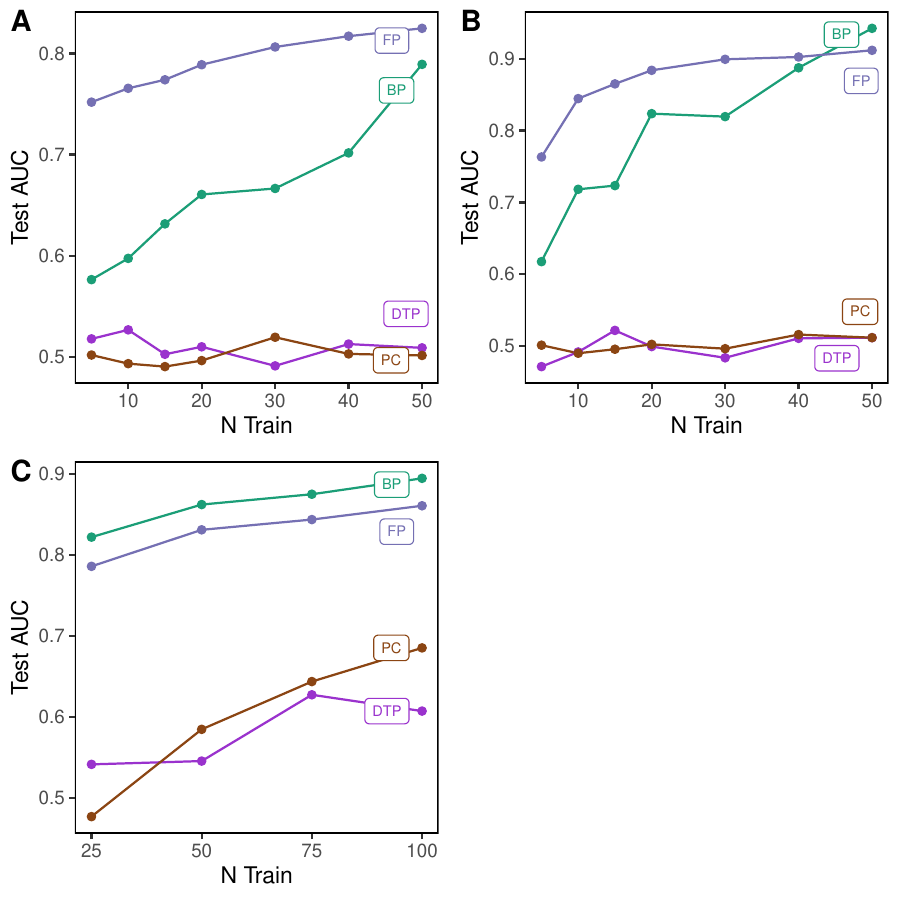}
   \caption{Test performance of few-shot trained 2D-CNN models, showing Difference Target Propagation (DTP) and Predictive Coding (PC) methods.  Mean test AUC is reported over 50 few-shot training experiments. A: Chest X-ray (CXR) task. B: Optical Coherence Tomography (OCT) task. C: CIFAR2 task in which models were required to classify the first two classes (aeroplane and automobile). Mean test AUC is reported over 50 few-shot training experiments. Models were fitted with $N\in\{5, 10,15, 20,30, 40, 50\}$ training samples from each class in CXR and OCT tasks and and $N\in \{25, 50,75, 100\}$ samples per class for the CIFAR2 task.}
  \label{fig: fewshot extra methods}

\end{figure}

Discrimination performance of Difference Target Propagation and Predictive Coding methods in few-shot learning tasks is plotted in Figure \ref{fig: fewshot extra methods}. Both methods proved uninformative for few-shot learning CXR (Figure \ref{fig: fewshot extra methods}-A), OCT (Figure \ref{fig: fewshot extra methods}-B), and CIFAR2 tasks (Figure \ref{fig: fewshot extra methods}-C).

Discrimination performance (AUC) of Forward Projection, random features, Local Supervision, Forward-Forward, Predictive Coding, Difference Target Propagation and  and backpropagation in few-shot modelling tasks is provided in Supplementary Table \ref{tab:fewshot_performance}.

\begin{table}
\tiny
\begin{tabular}{p{0.5cm}p{0.6cm}p{0.65cm}llllllll}
\toprule
 &  &  & \multicolumn{7}{r}{AUC} \\
 &  & Method & FP (ours) & RF & LS & FF & PC & DTP & BP \\
Dataset & Partition & $N_{\text{train}}$ &  &  &  &  &  &  &  \\
\midrule
\multirow[t]{14}{*}{CXR} & \multirow[t]{7}{*}{Train} & N=5 & 100.0 ± 0.0 & 100.0 ± 0.0 & 50.7 ± 16.0 & 52.0 ± 9.5 & 51.0 ± 12.9 & 53.9 ± 12.7 & 72.5 ± 9.4 \\
 &  & N=10 & 100.0 ± 0.0 & 100.0 ± 0.0 & 48.7 ± 12.7 & 48.8 ± 6.3 & 49.4 ± 9.9 & 53.8 ± 10.4 & 69.0 ± 8.1 \\
 &  & N=15 & 100.0 ± 0.0 & 100.0 ± 0.0 & 55.2 ± 13.9 & 51.1 ± 6.1 & 49.4 ± 9.5 & 50.4 ± 8.8 & 72.4 ± 10.6 \\
 &  & N=20 & 100.0 ± 0.0 & 100.0 ± 0.2 & 58.5 ± 12.8 & 50.0 ± 5.5 & 50.2 ± 9.6 & 52.0 ± 7.4 & 72.6 ± 8.0 \\
 &  & N=30 & 100.0 ± 0.0 & 99.9 ± 0.4 & 59.8 ± 15.5 & 50.0 ± 5.7 & 52.4 ± 9.0 & 49.4 ± 8.4 & 71.8 ± 9.6 \\
 &  & N=40 & 100.0 ± 0.1 & 99.3 ± 0.8 & 61.4 ± 15.8 & 49.7 ± 4.8 & 52.1 ± 8.9 & 52.3 ± 7.9 & 74.3 ± 10.2 \\
 &  & N=50 & 99.9 ± 0.2 & 98.4 ± 1.0 & 66.1 ± 14.8 & 49.4 ± 6.2 & 51.4 ± 8.2 & 52.2 ± 7.3 & 84.2 ± 9.8 \\

 & \multirow[t]{7}{*}{Test} & N=5 & 75.2 ± 7.0 & 67.6 ± 7.5 & 49.7 ± 8.8 & 49.3 ± 5.7 & 50.2 ± 8.1 & 51.8 ± 7.6 & 57.6 ± 7.6 \\
 &  & N=10 & 76.6 ± 5.8 & 68.1 ± 6.0 & 47.9 ± 9.0 & 49.1 ± 5.7 & 49.3 ± 8.9 & 52.7 ± 8.3 & 59.7 ± 6.2 \\
 &  & N=15 & 77.4 ± 4.5 & 69.2 ± 4.8 & 53.9 ± 12.3 & 49.9 ± 6.7 & 49.0 ± 8.3 & 50.2 ± 7.6 & 63.1 ± 8.6 \\
 &  & N=20 & 78.9 ± 3.7 & 69.0 ± 4.9 & 58.9 ± 11.5 & 49.0 ± 5.8 & 49.6 ± 8.5 & 51.0 ± 7.1 & 66.1 ± 9.3 \\
 &  & N=30 & 80.6 ± 3.7 & 69.5 ± 3.7 & 59.2 ± 14.0 & 51.2 ± 6.2 & 51.9 ± 8.8 & 49.1 ± 8.9 & 66.6 ± 8.0 \\
 &  & N=40 & 81.7 ± 2.9 & 72.4 ± 3.7 & 61.2 ± 16.1 & 48.0 ± 6.4 & 50.3 ± 7.6 & 51.3 ± 8.6 & 70.2 ± 8.3 \\
 &  & N=50 & 82.5 ± 3.5 & 72.2 ± 3.1 & 64.8 ± 14.9 & 49.4 ± 7.4 & 50.1 ± 7.9 & 50.9 ± 7.2 & 78.9 ± 8.3 \\

\multirow[t]{14}{*}{OCT} & \multirow[t]{7}{*}{Train} & N=5 & 100.0 ± 0.0 & 100.0 ± 0.0 & 53.9 ± 27.2 & 49.4 ± 12.5 & 52.5 ± 23.0 & 56.9 ± 21.5 & 86.8 ± 13.2 \\
 &  & N=10 & 100.0 ± 0.0 & 100.0 ± 0.0 & 58.3 ± 17.8 & 50.4 ± 12.2 & 51.4 ± 15.9 & 54.4 ± 13.8 & 83.3 ± 12.2 \\
 &  & N=15 & 100.0 ± 0.0 & 100.0 ± 0.0 & 57.1 ± 15.7 & 49.9 ± 11.2 & 49.6 ± 12.3 & 52.5 ± 13.8 & 80.5 ± 11.2 \\
 &  & N=20 & 100.0 ± 0.0 & 100.0 ± 0.0 & 53.4 ± 21.0 & 51.6 ± 9.8 & 52.9 ± 13.0 & 52.8 ± 11.7 & 88.9 ± 12.1 \\
 &  & N=30 & 100.0 ± 0.0 & 100.0 ± 0.0 & 58.0 ± 18.8 & 50.1 ± 11.3 & 50.1 ± 11.5 & 51.7 ± 9.7 & 83.7 ± 12.3 \\
 &  & N=40 & 100.0 ± 0.0 & 99.9 ± 0.3 & 61.4 ± 21.0 & 52.3 ± 9.2 & 52.2 ± 11.2 & 51.0 ± 12.6 & 88.0 ± 12.0 \\
 &  & N=50 & 100.0 ± 0.0 & 99.7 ± 0.7 & 61.7 ± 19.8 & 50.8 ± 13.6 & 51.1 ± 8.8 & 53.6 ± 11.0 & 95.2 ± 8.3 \\

 & \multirow[t]{7}{*}{Test} & N=5 & 76.3 ± 11.5 & 63.1 ± 10.0 & 51.1 ± 17.6 & 52.5 ± 12.0 & 50.1 ± 15.6 & 47.1 ± 10.5 & 61.8 ± 17.4 \\
 &  & N=10 & 84.5 ± 5.8 & 66.5 ± 7.7 & 54.6 ± 16.7 & 51.8 ± 12.1 & 49.0 ± 17.1 & 49.2 ± 11.6 & 71.8 ± 15.9 \\
 &  & N=15 & 86.5 ± 7.8 & 68.7 ± 7.4 & 56.8 ± 16.3 & 47.8 ± 12.8 & 49.6 ± 13.6 & 52.2 ± 12.9 & 72.3 ± 14.4 \\
 &  & N=20 & 88.4 ± 5.9 & 69.5 ± 6.0 & 49.6 ± 19.1 & 52.3 ± 13.0 & 50.2 ± 15.5 & 49.9 ± 14.7 & 82.4 ± 14.4 \\
 &  & N=30 & 89.9 ± 4.0 & 70.5 ± 5.7 & 55.7 ± 20.9 & 51.8 ± 11.7 & 49.6 ± 15.5 & 48.4 ± 13.9 & 81.9 ± 12.1 \\
 &  & N=40 & 90.2 ± 5.0 & 73.3 ± 5.5 & 62.0 ± 25.1 & 50.6 ± 13.0 & 51.6 ± 16.5 & 51.1 ± 15.4 & 88.7 ± 9.3 \\
 &  & N=50 & 91.2 ± 4.7 & 73.6 ± 5.9 & 62.6 ± 23.8 & 49.7 ± 18.4 & 51.2 ± 12.9 & 51.2 ± 13.4 & 94.2 ± 6.6 \\

\multirow[t]{8}{*}{CIFAR} & \multirow[t]{4}{*}{Train} & N=25 & 95.3 ± 2.7 & 100.0 ± 0.0 & 71.6 ± 23.6 & 52.6 ± 9.7 & 49.6 ± 17.5 & 55.9 ± 16.2 & 96.8 ± 4.5 \\
 &  & N=50 & 96.0 ± 2.0 & 100.0 ± 0.0 & 78.9 ± 18.3 & 59.2 ± 21.0 & 60.3 ± 16.5 & 56.1 ± 13.8 & 99.4 ± 1.9 \\
 &  & N=75 & 96.0 ± 1.6 & 100.0 ± 0.0 & 86.7 ± 5.2 & 67.5 ± 17.8 & 64.6 ± 15.1 & 63.4 ± 11.4 & 98.2 ± 2.5 \\
 &  & N=100 & 95.9 ± 1.3 & 100.0 ± 0.1 & 88.7 ± 3.8 & 73.7 ± 10.6 & 69.6 ± 11.0 & 60.9 ± 8.9 & 99.9 ± 0.3 \\

 & \multirow[t]{4}{*}{Test} & N=25 & 78.6 ± 3.4 & 78.6 ± 3.6 & 65.0 ± 20.5 & 52.5 ± 8.8 & 47.7 ± 14.6 & 54.2 ± 14.1 & 82.2 ± 2.4 \\
 &  & N=50 & 83.1 ± 2.8 & 77.3 ± 3.1 & 74.2 ± 16.5 & 60.1 ± 22.8 & 58.5 ± 16.3 & 54.6 ± 13.4 & 86.3 ± 2.2 \\
 &  & N=75 & 84.4 ± 1.7 & 76.2 ± 3.1 & 81.5 ± 2.6 & 68.3 ± 18.8 & 64.4 ± 15.0 & 62.8 ± 10.7 & 87.5 ± 2.1 \\
 &  & N=100 & 86.1 ± 2.0 & 78.1 ± 2.6 & 83.5 ± 2.8 & 74.3 ± 9.7 & 68.6 ± 11.9 & 60.7 ± 9.4 & 89.5 ± 1.4 \\
\bottomrule
\end{tabular}
\caption{Train and Test AUC performance in few-shot learning experiments. AUC: area under curve. CXR: chest x-ray dataset. OCT: optical coherence tomography dataset. BP: Backpropagation; DTP: Difference Target Propagation; FF: Forward Forward; FP: Forward Projection; LS: Local Supervision; PC: Predictive Coding; RF: Random features}
\label{tab:fewshot_performance}
\end{table}

\subsection{Training Times}\label{remark:training times}
Training times and epochs for each method, in full dataset modelling. FMNIST was modelled with a $3\times 1000$ neuron MLP network. Sequential datasets (PTBXL-MI and Promoters) were modelled by a 1D-CNN architecture of four convolutional blocks with $32 \times 2^{l-1}$ filters in the $l$-th block. CIFAR2 modelling used a vision transformer architecture operating on image patches of dimension $4\times4$, with a sequential stack of four multi-headed attention layers, each having 8 heads, embedding dimension 64, and MLP dimension 64. All experiments were run on the Google Colab service using an NVIDIA T4 graphics processing unit.

\begin{table}
\tiny
\begin{tabular}{llllllllll}
\toprule
 &  & Method & FP & RF & LS & FF & PC & DTP & BP \\
 & Dataset & Metric &  &  &  &  &  &  &  \\
\midrule
\multirow[t]{8}{*}{} & FMNIST & Time (s)& 0.3 ± 0.1 & 0.1 ± 0.0 & 46.9 ± 12.5 & 171.9 ± 56.6 & 150.2 ± 61.2 & 72.7 ± 26.4 & 22.4 ± 4.8 \\

 & Promoters & & 6.5 ± 0.2 & 3.3 ± 0.6 & 91.6 ± 37.7 & 69.5 ± 12.4 & 258.9 ± 233.7 & 103.0 ± 23.6 & 30.9 ± 4.3 \\

 & PTBXL-MI & & 10.6 ± 1.1 & 6.7 ± 0.8 & 56.3 ± 13.8 & 35.5 ± 11.9 & 73.3 ± 15.8 & 61.0 ± 24.2 & 18.7 ± 5.1 \\

 & CIFAR2& & 16.3 ± 0.6 & 9.2 ± 0.6 &  &  &  &  & 43.3 ± 15.1 \\

 & FMNIST & Epochs& 1.0 ± 0.0 & 1.0 ± 0.0 & 10.8 ± 3.6 & 40.2 ± 13.8 & 25.4 ± 10.2 & 13.6 ± 5.3 & 11.6 ± 2.7 \\

 & Promoters & & 1.0 ± 0.0 & 1.0 ± 0.0 & 11.4 ± 4.9 & 6.2 ± 1.3 & 18.4 ± 17.5 & 7.8 ± 2.0 & 10.0 ± 1.4 \\

 & PTBXL-MI & & 1.0 ± 0.0 & 1.0 ± 0.0 & 17.0 ± 4.3 & 7.2 ± 2.3 & 14.6 ± 3.3 & 13.8 ± 5.9 & 13.2 ± 3.6 \\

 & CIFAR & & 1.0 ± 0.0 & 1.0 ± 0.0 &  &  &  &  & 10.8 ± 3.9 \\

\bottomrule
\end{tabular}
\caption{Training times and epochs in full dataset modelling experiments. BP: Backpropagation; DTP: Difference Target Propagation; FF: Forward Forward; FP: Forward Projection; LS: Local Supervision; PC: Predictive Coding; RF: Random features}
\label{tab:timing performance}
\end{table}

\end{appendices}

\end{document}